%% file: acl_latex.tex
\pdfoutput=1
\PassOptionsToPackage{table}{xcolor}

\documentclass[11pt]{article}

\usepackage[final]{acl}

\usepackage{times}
\usepackage{latexsym}
\usepackage{bm}
\usepackage{listings}

\usepackage[T1]{fontenc}

\usepackage[utf8]{inputenc}
\usepackage{caption}  
\usepackage{booktabs} 

\usepackage{microtype}

\usepackage{inconsolata}
\usepackage{amsmath}

\usepackage{graphicx}

\usepackage{graphicx}
\usepackage{amssymb}

\usepackage{tabularray} 
\usepackage{makecell}

\usepackage{algorithm}
\usepackage{algpseudocode}

\usepackage{booktabs}
\usepackage{amsmath}
\usepackage[many]{tcolorbox}

\title{LeanK: Learnable K Cache Channel Pruning for Efficient Decoding}

\author{
  Yike Zhang$^{1}$\thanks{Work during internship at Microsoft.}, Zhiyuan He$^2$, Huiqiang Jiang$^2$, Chengruidong Zhang$^2$,  \\ \textbf{Yuqing Yang$^2$, Jianyong Wang$^1$, Lili Qiu$^2$} \\
  $^1$Tsinghua University, $^2$Microsoft Research \\
  \tt zhangyik21@mails.tsinghua.edu.cn, zhiyuhe@microsoft.com
}

\begin{document}
\maketitle
\begin{abstract}
Large language models (LLMs) enable long-context tasks but face efficiency challenges due to the growing key-value (KV) cache. We propose LeanK, a learning-based method that prunes unimportant key (K) cache channels by leveraging static channel sparsity. With a novel two-stage training process, LeanK learns channel-wise static mask that could satisfy specific sparsity ratio and hardware alignment requirement. LeanK reduces GPU memory and accelerates decoding without sacrificing accuracy. Experiments demonstrate up to 70\% K cache and 16\%-18\% V cache memory reduction. Custom decoding kernel enables 1.3x speedup for attention computation. We also provide insights into model channels and attention heads during long-context inference by analyzing the learned importance distribution. Our code is available at \href{https://aka.ms/LeanK}{https://aka.ms/LeanK}. 
\end{abstract}

\input{sections/intro_ver2}

\input{sections/motivation}

\input{sections/method}

\input{sections/experiments}

\input{sections/analysis}

\section{Related Works}

\noindent \textbf{KV Cache Optimization.} Large KV caches in long-context LLMs lead to significant GPU memory overhead and increasing output latency. To mitigate this, various optimization methods have been proposed. Eviction-based methods discard KV entries of less important tokens, such as H2O \cite{zhang2023h2o} and SnapKV \cite{li2024snapkv}, which rely on attention scores, or DuoAttention \cite{xiao2024duoattentionefficientlongcontextllm}, which prunes KV cache at the head level. Selection-based methods like SparQ \cite{ribar2023sparq}, Quest \cite{tang2024questqueryawaresparsityefficient}, and Double Sparsity \cite{yang2024posttrainingsparseattentiondouble} retain the full KV cache in memory but selectively read relevant entries to reduce memory bandwidth usage. Quantization methods such as KIVI \cite{liu2024kivi} compress KV caches by reducing numerical precision. Our method is orthogonal to these eviction, selection, and quantization approaches and can be combined with them for further gains (see~\S\ref{sec:orthogonal}).

\noindent \textbf{Structured Pruning.} Traditional structured pruning methods for LLMs target hidden states \cite{ma2023llm}, layers \cite{gromov2403unreasonable}, and expert components \cite{lu2024not}, but they are limited to small task ranges and often suffer from poor performance.  
Other pruning methods primarily target weights and activations \cite{frantar2023sparsegptmassivelanguagemodels, sun2024simpleeffectivepruningapproach}. In contrast, our approach focuses on pruning the KV cache and attention computations, incorporating the unstructured attention sink and local window mechanism into algorithm design, which serves as a valuable complement to existing weight pruning methods for LLMs.

\input{sections/conclusion}

\section*{Limitations}

We observe significant redundancy along the channel dimension of pretrained LLMs. Improving positional embeddings and conducting more thorough pretraining over this dimension may enhance the model's long-context processing ability and reduce memory consumption. We leave this exploration for future work.

\bibliography{custom}

\input{sections/appendix}

\end{document}

%% file: sections/intro_ver2.tex
\section{Introduction}

Large language models (LLMs) have advanced to support long-context tasks such as document understanding \cite{li-etal-2024-loogle}, multi-turn dialogues \cite{yi2024survey}, repository-level code completion \cite{jimenez2024swebench}, and complex reasoning \cite{zhou2025gsminfinitellmsbehaveinfinitely}. However, efficient inference under these long-context settings remains challenging due to the growing size of the key-value (KV) cache, which not only significantly increases GPU memory usage but also repeatedly stresses GPU memory bandwidth during token generation, leading to slower inference speeds \cite{zhang2023h2o,tang2024questqueryawaresparsityefficient,liu2024kivi}.

Existing efforts to optimize the KV cache include: (1) Eviction, which discards cache of less important tokens \cite{li2024snapkv, zhang2023h2o} or cache in less important attention heads \cite{xiao2024efficientstreaminglanguagemodels, xiao2024duoattentionefficientlongcontextllm}; (2) Selection, which retains the full KV cache but selectively reads relevant entries during inference \cite{tang2024questqueryawaresparsityefficient, chen2024magicpiglshsamplingefficient, liu2024retrievalattentionacceleratinglongcontextllm}; and (3) Quantization, which compresses the KV cache using compressed data types \cite{liu2024kivi}. 
Despite the effectiveness of these methods, they typically assume that all channels in the key (K) cache are equally necessary when the final attention score is calculated, which limits their efficiency optimization potential.

We identify a unique and largely underexplored opportunity for optimizing the K cache by leveraging the sparsity in its channel dimension. Specifically, we observe: 
(1) Previous studies suggest that RoPE influences the feature encoded in each dimension of K \cite{barbero2025roundroundgomakes}. Dimensions associated with high frequencies tend to be less stable for text retrieval, revealing a potential opportunity for pruning . (2) Besides, we find that the importance of K cache channels tends to be static and can be determined offline. This static sparsity can offer consistent speedups during online inference. 
(3) K channel sparsity is orthogonal to existing approaches, and can be combined with them for further acceleration.

Based on our observation, we propose LeanK, a learning-based method for pruning the channel dimension of the K cache to enable efficient long-context decoding. LeanK learns static sparsity through a double-stage process. In the first stage, it estimates the global importance of each K channel. In the second stage, it learns a sparse channel mask that adheres to a target sparsity ratio and is optimized for hardware efficiency. At inference time, LeanK prunes the K cache channels based on the learned static sparsity, significantly reducing GPU memory usage and improving decoding speed. 

\begin{figure*}[ht]
    \centering
    \begin{minipage}[t]{0.3\textwidth}
    \vspace{-\baselineskip}
        \includegraphics[width=\textwidth]{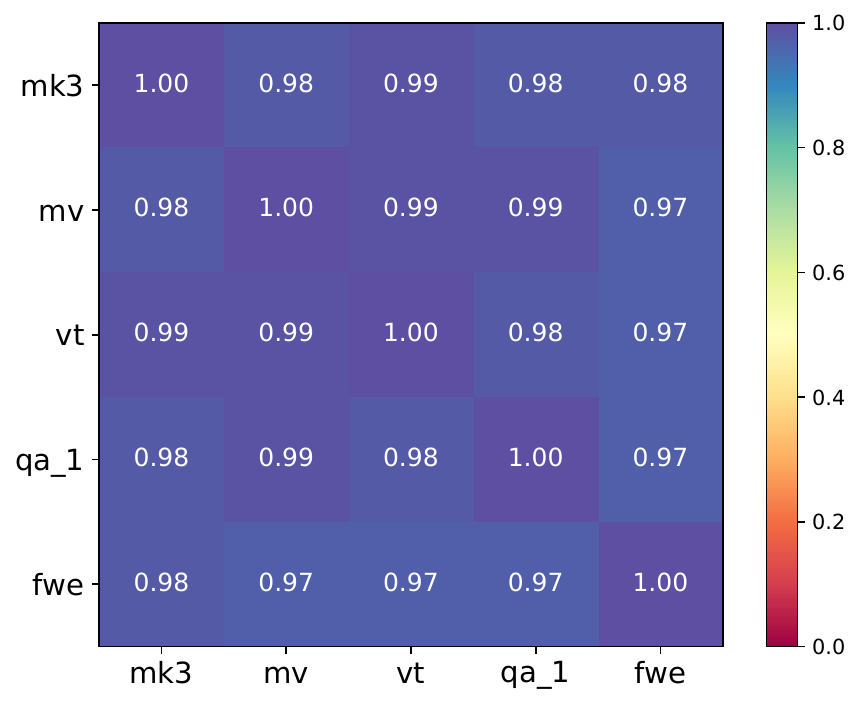} 
        \label{fig:pearson}
    \end{minipage}\hfill
    \begin{minipage}[t]{0.3\textwidth}
    \vspace{-\baselineskip}
        \includegraphics[width=\textwidth]{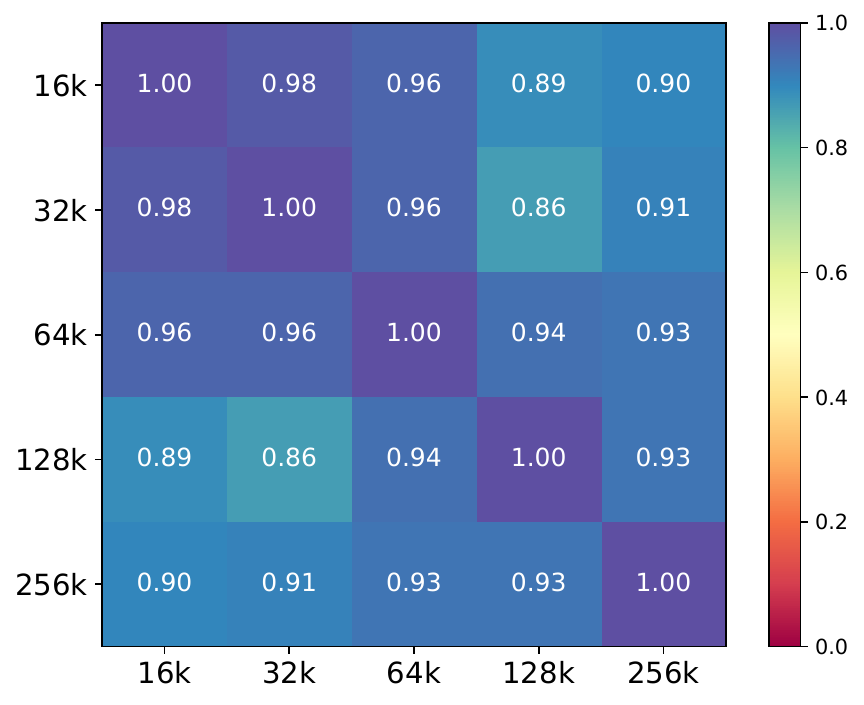} 
        \label{fig:pearson2}
    \end{minipage}\hfill
    \begin{minipage}[t]{0.3\textwidth}
    \vspace{-\baselineskip}
        \includegraphics[width=\textwidth]{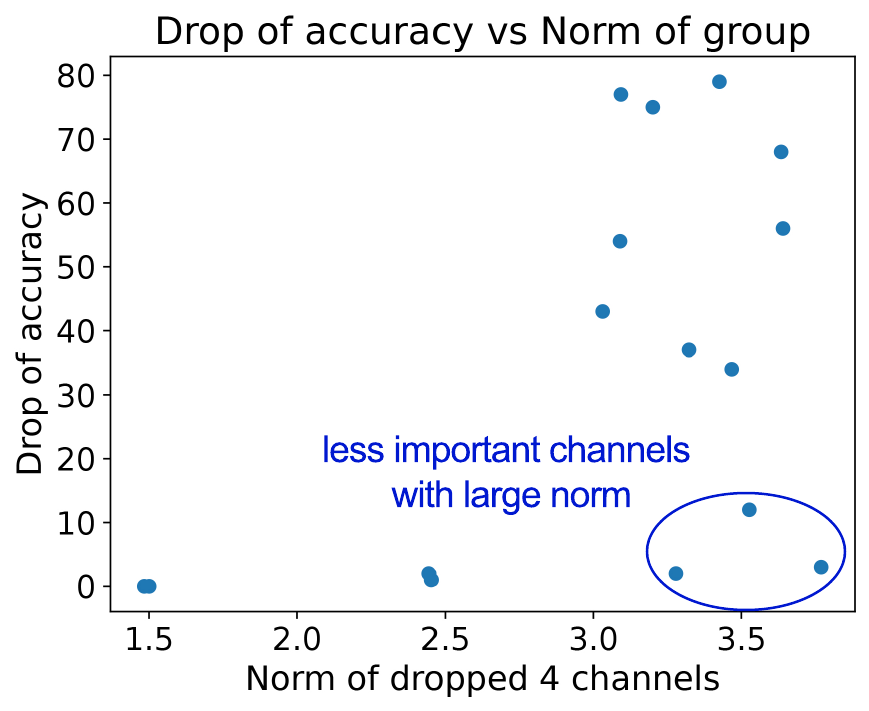}
        \label{fig:drop_acc}
    \end{minipage} 
    \vspace{-12pt}
     \caption{
     \textbf{Left} and \textbf{Middle}: K channel sparsity remains static across different RULER \textbf{tasks} and \textbf{sequence lengths}.
     \textbf{Right}: There exist channels with relatively \textbf{large norm} but has \textbf{limited impact} on end-to-end performance. 
     }
    \vspace{-12pt}
    \label{fig:motivation}
\end{figure*}

We conduct extensive experiments on two recent long-context LLMs, Llama-3.1-8B-Instruct \cite{grattafiori2024llama3herdmodels} and Qwen2.5-7B-Instruct \cite{qwen2025qwen25technicalreport}, across three different benchmarks. Results show that LeanK enables approximately 70\% GPU memory reduction in K cache and 16\%--18\% memory reduction in V cache size. 
Custom decoding kernel enables 1.3x speedup for attention computation while preserving model accuracy. 
Moreover, LeanK is highly compatible with existing KV cache optimization techniques. When combined with quantization methods such as KIVI \cite{liu2024kivi}, LeanK improves the overall KV cache compression ratio from 5.3$\times$ to 9.7$\times$, substantially alleviating memory bottlenecks in long-context inference. 
Furthermore, we analyze the learned channel-wise importance distribution of K cache and gained insights into model's behavior related to RoPE.

%% file: sections/motivation.tex
\section{Motivation}
\label{sec:motiv}

The motivation of our method is based on the following three observations.

\subsection{The use of RoPE introduces channel inefficiency in K.}
In modern LLMs, RoPE is applied to both Q and K in Transformer attention. RoPE encodes positional information by assigning each pair of K dimensions a specific frequency. However, recent studies show that high-frequency dimensions tend to be unstable and contribute little to long-context inference \cite{hong-etal-2024-token, 
barbero2024round}. These findings indicate that many K channels are underutilized during long-context inference, presenting an opportunity for effective pruning.

\subsection{Sparsity in K channels tends to be static.}

We assess the staticity of important K channels by computing Pearson correlation coefficient \cite{sedgwick2012pearson} between channel norm distribution of different input sequences.\footnote{Detailed methods of computing channel norm distribution is provided in Appendix~\ref{appendix:k-channel-staticity}, which roughly assesses each channel's importance through its norm. } As shown in Figure~\ref{fig:motivation}, channel norm distribution on Llama-3.1-8B-Instruct exhibits consistently high Pearson correlation coefficients across five diverse RULER tasks~\cite{hsieh2024rulerwhatsrealcontext} and multiple sequence lengths, suggesting an inherent staticity in channel importance.

In contrast, ThinK \cite{xu2025thinkthinnerkeycache} assumes dynamic sparsity in K channels, estimating each channel's importance dynamically via $\bm{Q}_{window}\bm{K}^T$ during inference, where $\bm{Q}_{window}$ corresponds to several recent context tokens. Instead, we test a simpler static, norm-based pruning approach. We derive a static pruning mask based on the average norm of each channel across 100 input sequences from RULER 64K NIAH\_multikey3 task, and apply this mask universally across all tasks. Results in Table \ref{tab:static} show that the static method achieves comparable performance to ThinK. 

\begin{table}[h]
    \centering
    \setlength{\tabcolsep}{1.8mm}
    \resizebox{1\columnwidth}{!}{
    \begin{tabular}{c|lc|c}
        \toprule
        & Method & Pruning Ratio & Acc \\
        \midrule
          & Original & - & 84.38 \\
        Llama-3.1-8B-Instruct & ThinK (Dynamic norm) & 60\% & 80.56 \\
        & Static norm & 60\% & 81.57 \\
        \bottomrule
    \end{tabular}
    }
    \vspace{-8pt}
    \caption{RULER 64K performance of static norm-based pattern and dynamic norm-based method (ThinK). Detailed results and analysis are in Appendix \ref{sec:static}.}
    \label{tab:static}
    \vspace{-16pt}
\end{table}

\subsection{Some channels exhibit large magnitudes but limited impact.}

Furthermore, We conduct experiments by removing subsequent groups of every 4 channels from Llama-3.1-8B-Instruct model and evaluate the resulting performance degradation on RULER 64K NIAH\_multikey3 task. Relationship between average norm of each group and corresponding performance drop is presented in Figure \ref{fig:motivation}. There exist channels with large norm but have little impact on model performance (marked by the blue circle). 
Relying solely on magnitude to decide channel importance may overlook the heterogeneity between different decoder layers and attention heads and miss some pruning opportunities.

%% file: sections/method.tex
\begin{figure*}[ht]
    \centering
    \includegraphics[width=1\linewidth]{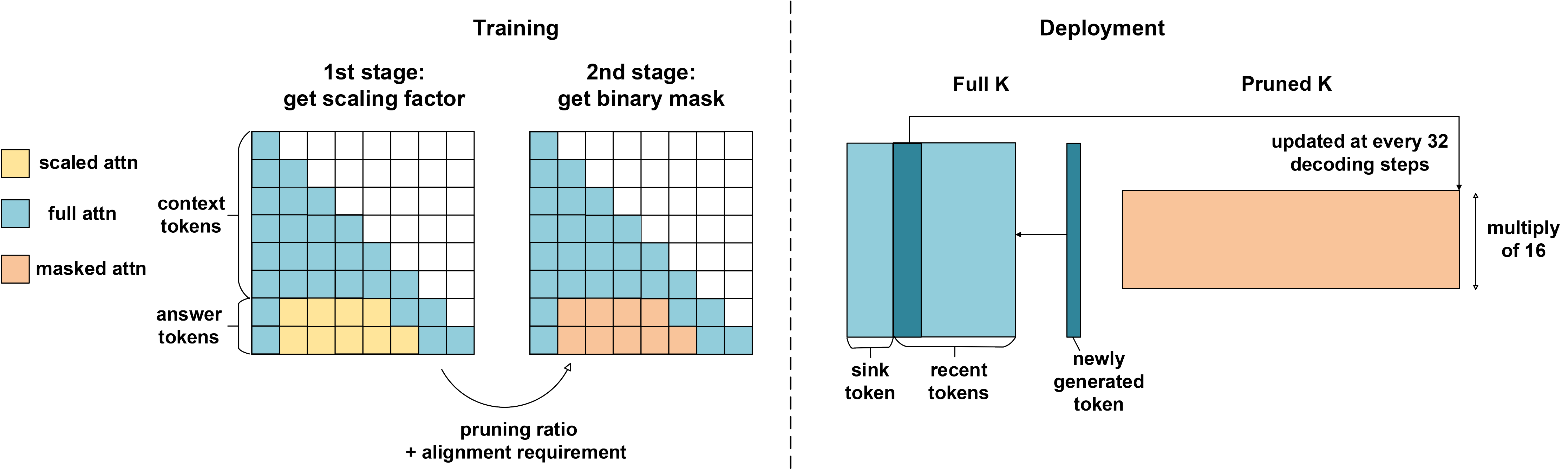}
    \vspace{-16pt}
    \caption{Overall demonstration of LeanK's double-stage training and deployment method. }
    \label{fig:method}
        \vspace{-8pt}
\end{figure*}

\section{Method}

Based on our observations, we present \textbf{LeanK}, a learning-based method that exploits channel sparsity in K cache to accelerate long-context decoding. 
Our goal is to learn a binary mask for pruning K channels according to a predefined pruning ratio, using a \textbf{double-stage process}. 
In the first stage, we learn a \textbf{continuous scaling factor} representing the global importance of each K channel. In the second stage, we convert the learned scaling factor into a \textbf{binary mask} suitable for deployment. 

\subsection{Training stage 1}

Consider a transformer model with $L$ layers, where each layer has either Multi-Head Attention (MHA) or Grouped Query Attention (GQA) containing $n$ heads (or groups), each with a dimension of $d$. We introduce a scaling factor $\bm{\alpha} \in \mathbb{R}^{L \times n \times d}$, initialized with all elements set to 1. 
The value of $\bm{\alpha}$ represents the global importance score of each channel and will be learned in the first stage.

Suppose the input sequences for learning are represented as $\bm{X} = [\bm{X}_{\text{ctx}};\bm{X}_{\text{ans}}]$, where the semicolon denotes concatenation. Each sequence consists of context tokens $\bm{X}_{\text{ctx}}$ and answer tokens $\bm{X}_{\text{ans}}$. Since the decoding phase is the primary focus of our work, the training loss is computed only based on the answer part. Initially, standard full attention is applied, from which we obtain the hidden states corresponding to the answer tokens from the last layer, denoted by $\bm{H}_{\text{full}} \in \mathbb{R}^{N_{\text{ans}} \times n \times d}$, where $N_{\text{ans}}$ is the number of answer tokens.

Then, we apply a specialized \textit{scaled attention} based on $\bm{\alpha}$. As shown in Figure \ref{fig:method}, the attention corresponding to $\bm{X}_{ctx}$ in each head is still full attention, formulated as:
\[{\bm{A}}_{\text{ctx}} = \mathrm{softmax}(\bm{Q}_{\text{ctx}}\bm{K}_{\text{ctx}}^{T} \odot \bm{M}_{\text{causal}}) \bm{V}\]

For \(\bm{X}_{\text{ans}}\), we scale the attention logit as:
\begin{equation}
\begin{aligned}
    \bm{L}_{\text{ans}} & = \bm{Q}_{\text{ans}}\bm{K}^{T} \odot {\bm{M}_{\text{s+l}}} \\
                 & \quad + \bm{Q}_{\text{ans}}(\bm{K}\mathrm{diag}(\bm{\alpha}_{i,j}))^T \odot {\bm{M}_{\text{mid}}} \\
\end{aligned}
\label{eq:att-logit}
\end{equation}
where $\bm{M}_{\text{s+l}}$ is a binary mask preserving only the sink and sliding window attention, while $\bm{M}_{\text{mid}}$ preserves only the middle attention region (excluding sink and sliding windows), as illustrated in Figure~\ref{fig:method}. The scaling factor $\bm{\alpha}_{i,j} \in \mathbb{R}^{d}$ corresponds to layer $i$, head $j$, and scales each channel dimension of the key matrix $\bm{K} \in \mathbb{R}^{N\times d}$ ($N$ denotes the number of input tokens), producing $\bm{K}\,\mathrm{diag}(\bm{\alpha}_{i,j})$. Due to masks $\bm{M}_{\text{s+l}}$ and $\bm{M}_{\text{mid}}$, scaling affects only the middle-region attention logits, keeping the sink and sliding window attention intact, as they are more critical and consume constant GPU memory regardless of sequence length ~\cite{xiao2024efficientstreaminglanguagemodels}. In contrast, the middle attention grows with the number of tokens, making it the primary target for pruning. Then, the scaled attention logits for the answer tokens are computed as:
\begin{equation*}
\begin{aligned}
    \bm{A}_{\text{ans}} & = \mathrm{softmax}(\bm{L}_{\text{ans}} \odot \bm{M}_{\text{causal}}) \bm{V}
\end{aligned}
\end{equation*}

In each attention head, the attention maps for context and answer tokens are concatenated as \(\bm{A} = [\bm{A}_{\text{ctx}}; \bm{A}_{\text{ans}}]\). We obtain the hidden states from the final layer corresponding to the answer tokens, denoted as \(\bm{H}_{\text{scaled}} \in \mathbb{R}^{N_{\text{ans}} \times n \times d}\).

The training loss for the first stage is defined as:
\begin{equation*}
\mathcal{L}_{\text{1st}} = \left\| \bm{H}_{\text{full}} - \bm{H}_{\text{scaled}} \right\|_2^2 + \lambda \left\| \bm{\alpha} \right\|_1,
\end{equation*}
where the first term is an L2 distillation loss that encourages the scaled hidden states to remain close to the full ones, and the second term is an L1 regularization that promotes sparsity in the learned scaling factors $\bm{\alpha}$. The coefficient $\lambda$ controls the trade-off between performance preservation and sparsity.

\begin{figure*}[t]
\centering
    \includegraphics[width=0.75\textwidth]{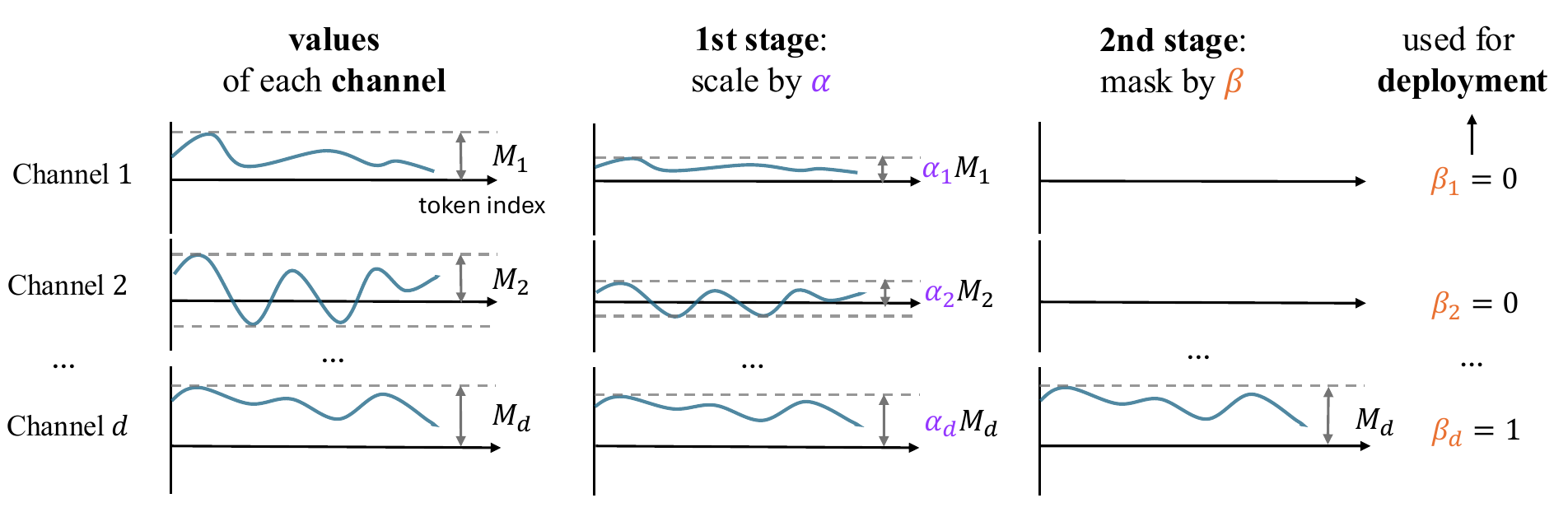} 
    \vspace{-8pt}
        \caption{Visualization of the training objectives of the two training stages. }
        \vspace{-14pt}
        \label{fig:vis_scaling_masking}
\end{figure*}

We train our model on two passkey retrieval tasks: (1) Dense retrieval, where $\bm{X}_{\text{ctx}}$ consists of key-value pairs and the goal is to retrieve the value for a given key as $\bm{X}_{\text{ans}}$; and (2) Multi-value retrieval, where $\bm{X}_{\text{ctx}}$ includes distraction text and keys with multiple values, and the task is to retrieve all values for a given key. All keys and values are randomly generated. We select these tasks because the first is challenging and thus effective for preserving the model's retrieval capabilities, while the second task involves generating relatively long answers, potentially improving the model's long-term generation performance.

\subsection{Training stage 2}
\label{sec:2ndstage}

The scaling factor $\bm{\alpha} \in \mathbb{R}^{L \times n \times d}$ learned in the first stage encodes the importance of each channel. Our next goal is to derive a channel-wise binary mask $\bm{\beta} \in \{0,1\}^{L \times n \times d}$ that could directly be used for pruning, where $\bm{\beta}_{i,j,k} = 0$ indicates that the $k$-th channel of the K cache in layer $i$ and head $j$ should be pruned. 
We require such a binary mask to satisfy two requirements: (\textbf{R1}) The desired \textbf{pruning ratio} \(s\%\) should be specified before deployment, and (\textbf{R2}) The mask should be GPU-friendly, i.e., the number of remained channels for each head should satisfy the \textbf{alignment requirement} for efficient memory loading and computation (e.g. be a multiply of \(r = 16\) or \(32\)).

Training stage 1 does not take these two requirements into consideration, bringing up the necessity of a second training stage. 
In the second training stage we optimize a binary mask \(\beta\) generated by:

\begin{equation*}
\bm{\beta} = \text{Top}_{s\%, r}(\bm{\alpha}),
\end{equation*}
 where \( \text{Top}_{s\%, r}(\cdot) \) is a two-step operator. First, \( \text{Top}_{s\%}(\bm{\alpha}) \) selects the top \( s\% \) most important channels across all heads. Then, for each layer \( i \) and head \( j \), let \( n_{i,j} \) be the number of channels initially selected. We round \( n_{i,j} \) to the nearest multiple of \( r \), denoted by \( n_{i,j}' = \left\lfloor \frac{n_{i,j}}{r} \right\rceil \times r \), and keep the top \( n_{i,j}' \) entries from \( \bm{\alpha}_{i,j} \) to form the final mask: \( \bm{\beta}_{i,j} = \text{Top}_{n_{i,j}'}(\bm{\alpha}_{i,j}) \).

The scaled attention in the second stage is applied similarly as in the first stage, but with $\bm{\alpha}$ replaced by the binary mask $\bm{\beta}$ in Equation~\ref{eq:att-logit}. Since the purpose of this stage is not to induce additional sparsity but to preserve model performance under pruning, the training objective only includes the distillation loss that aligns the model performance:
\[
\mathcal{L}_{2nd} =  \left\| \bm{H}_{\text{full}} - \bm{H}_{\text{scaled}} \right\|_2^2.
\]

Figure~\ref{fig:vis_scaling_masking} illustrates the roles of the scaling factor \(\alpha\) and binary mask \(\beta\) in the double-stage training process. \textbf{Both stages are crucial} for learning an effective pruning mask. Directly applying Top-K to \(\alpha\) results in suboptimal performance as validated in 
Appendix~\ref{sec:stage2_detail} (relying on test-time to decide pruning ratio misaligns with \textbf{R1}), and might be unfavorable for GPU efficiency (violates \textbf{R2}). Conversely, skipping stage 1 and directly optimizing a binary mask is difficult, especially at high pruning ratios. In practice, we observe that such training approach often fails to converge.

\subsection{Deployment}
\label{sec:deployment}

During deployment, after the prefilling stage, the key cache $\bm{K}$ is pruned and partitioned into two parts, $\bm{K} = [\bm{K}_{\text{s+l}}; \bm{K}_{\text{prun}}]$, where $\bm{K}_{\text{s+l}}$ includes the full K cache for attention sink and local windows, and $\bm{K}_{\text{prun}}$ is the cache pruned based on the binary mask $\bm{\beta}$. As illustrated in Figure~\ref{fig:method}, during decoding, the oldest key vector in the local window is taken from $\bm{K}_{\text{s+l}}$, pruned using $\bm{\beta}$, and appended to $\bm{K}_{\text{prun}}$; the key of the newly generated token is then appended to $\bm{K}_{\text{s+l}}$. To reduce overhead, we perform this update every 32 tokens instead of every step. Attention computation at each decoding step is:
\begin{equation*}
\bm{A}= \mathrm{softmax}(\bm{q}\bm{K}_{\text{s+l}}^T + \bm{q}_{\text{prun}}\bm{K}_{\text{prun}}^T)\bm{V},
\end{equation*}
where $\bm{q}_{\text{prun}}$ represents the query vector pruned in each attention head according to $\bm{\beta}$. Notably, some attention heads have all channels pruned, making $\bm{K}_{\text{prun}}$ and its associated value cache unnecessary. For these heads, the attention calculation simplifies to:
$$
\bm{A}= \mathrm{softmax}(\bm{q}\bm{K}_{\text{s+l}}^T)\bm{V}_{\text{s+l}},
$$
where $\bm{V}_{\text{s+l}}$ corresponds exclusively to the sink and local window tokens, allowing additional memory savings in the V cache.

%% file: sections/experiments.tex
\begin{table*}[ht]
\setlength{\tabcolsep}{3pt}
\resizebox{\textwidth}{!}{
\begin{tabular}{lc|cccccccccccccc}
\hline
\textbf{Methods}                  & \textbf{Avg.} & \rotatebox{45}{\textbf{2Wiki}} & \rotatebox{45}{\textbf{GovRep}} & \rotatebox{45}{\textbf{Hotpot}} & \rotatebox{45}{\textbf{LCC}} & \rotatebox{45}{\textbf{MNews}} & \rotatebox{45}{\textbf{MF-en}} & \rotatebox{45}{\textbf{PsgCnt}} & \rotatebox{45}{\textbf{PsgRtr}} & \rotatebox{45}{\textbf{Qasper}} & \rotatebox{45}{\textbf{Rbench}} & \rotatebox{45}{\textbf{Samsum}} & \rotatebox{45}{\textbf{TREC}} & \rotatebox{45}{\textbf{TrvQA}} & \rotatebox{45}{\textbf{QMSum} } \\ \hline
\textit{Llama-3.1-8B-Instruct}  & 52.4  & 48.5  & 34.5  & 58.1 & 63.4  & 27.1  & 56.6  & 9.3  & 99.5  & 44.7  & 56.7  & 43.8  & 73.0  & 91.7 & 27.1 \\
ThinK 60\%   &  51.5 & 48.1 & 30.3 & 57.9 & 63.1 & 25.3 & \textbf{57.6} & \textbf{9.9} & \textbf{99.5} & 44.9 & 55.5 & 41.1 & 72.0 & 90.1 & \textbf{25.2} \\
ThinK 70\%   &  49.4 & 47.1 & 27.0 & 56.9 & 62.7 & 25.9 & 50.5 & 9.3 & \textbf{99.5} & 35.9 & 54.9 & 42.2 & 65.0 & 90.5 & 23.7 \\
\rowcolor{gray!20} LeanK 70\%   &  \textbf{52.2} & \textbf{48.3} & \textbf{33.3} & \textbf{58.1} & \textbf{63.2} & \textbf{26.5} & 56.3 & 9.6 & \textbf{99.5} & \textbf{46.5} & \textbf{57.0} & \textbf{42.7} & \textbf{73.0} & \textbf{92.1} & 25.1 \\ \hline

\textit{Qwen2.5-7B-Instruct} & 51.7 & 47.1 & 31.8 & 57.7 & 60.6 & 23.9 & 52.6 & 8.5 & 100.0 & 43.6 & 66.8 & 46.2 & 71.5 & 89.3 & 23.8 \\
ThinK 70\%    & 49.2 & 44.9 & 28.7 & 54.2 & \textbf{59.8} & 23.1 & 49.9 & 8.5 & 99.0 & 37.2 & 64.5 & \textbf{44.0} & 65.5 & 87.5 & 22.7 \\
\rowcolor{gray!20} LeanK 70\%    & \textbf{50.1} & \textbf{46.7} & \textbf{30.7} & \textbf{57.7} & 59.1 & \textbf{23.9} & \textbf{52.1} & \textbf{9.0} & \textbf{100.0} & \textbf{42.1} & \textbf{65.8} & 43.9 & \textbf{71.5} & \textbf{87.9} & \textbf{22.8}  \\ \hline
\end{tabular}}
\vspace{-5pt}
\caption{Performance of different methods and models on LongBench.}
\vspace{-8pt}
\label{tab:longbench_rst}
\end{table*}

\section{Experiments}

\subsection{Settings}

\noindent \textbf{LLMs.} We experiment with two recent and widely-used LLMs, Llama-3.1-8B-Instruct \cite{grattafiori2024llama3herdmodels} and Qwen2.5-7B-Instruct \cite{qwen2025qwen25technicalreport}, both supporting a 128K context window.

\noindent \textbf{Baselines.}  We mainly compare with ThinK \cite{xu2025thinkthinnerkeycache}, which uses a dynamic, query-driven strategy to prune unimportant channels in each attention head. It estimates channel importance via query–key multiplication at the end of prefilling. To our knowledge, it is the only method targeting pruning along the K channel dimension as we do. We also compare our channel selection method with Double Sparsity \cite{yang2024posttrainingsparseattentiondouble}, with results discussed in Appendix~\ref{sec:ds_comparison}.

\noindent \textbf{Benchmarks.} We evaluate our method on three long-context benchmarks: (1) LongBench \cite{bai2024longbenchbilingualmultitaskbenchmark}, a realistic and diverse benchmark where we evaluate on all tasks, including QA, few-shot learning, code generation, summarization, and counting; (2) RULER \cite{hsieh2024rulerwhatsrealcontext}, a challenging benchmark with tasks such as hay-in-the-stack, KV retrieval, variable tracking, and QA, evaluated at input lengths from 4K to 128K with 200 samples per task; (3) GSM-Infinite \cite{zhou2025gsminfinitellmsbehaveinfinitely}, a benchmark designed to assess mathematical reasoning under long-generation settings, where we evaluate on Medium and Hard tasks with context lengths of 8K, 16K and 32K.

\noindent \textbf{Implementation Details.} We apply the training method to Llama-3.1-8B-Instruct and Qwen2.5-7B-Instruct. Since Qwen2.5 adopts Yarn \cite{peng2023yarn} for context lengths beyond 32K, we train separate pruning masks for $\leq$32K and $>$32K contexts.  Training token lengths are uniformly sampled from 16K–96K for Llama, 4K–32K for original Qwen, and 64K–96K for Qwen with Yarn. The double-stage training consists of 2000 steps in the first stage and 200 steps in the second, with the latter using half the learning rate for stability. Specifically, we use learning rates of 0.02/0.01 for Llama and 0.04/0.02 for Qwen. We set $\lambda=0.06$ for both models, use a local window size of 1024, an attention sink size of 128, and optimize scaling factors using the Adam optimizer~\cite{kingma2017adammethodstochasticoptimization}.

\begin{table}[H]
\centering
\setlength{\tabcolsep}{1.8mm}
\resizebox{\columnwidth}{!}{
\begin{tabular}{l|ccccccc}
\hline
\textbf{Methods}      & \textbf{4K} & \textbf{8K} & \textbf{16K} & \textbf{32K} & \textbf{64K} & \textbf{128K} & \textbf{Avg.} \\ \hline
\textit{Llama-3.1-8B} & 95.1        & 93.1        & 90.2         & 86.0         & 84.4         & 73.5          & 87.1          \\
ThinK, 60\%           & 89.5        & 81.7        & 79.7         & 81.8         & 80.6         & 69.8          & 80.5          \\
ThinK, 70\%           & 58.3        & 39.2        & 37.5         & 36.4         & 35.3         & 40.0          & 41.1          \\
\rowcolor{gray!20} LeanK, 70\%           & \textbf{95.3}        & \textbf{93.4}        & \textbf{88.8}         & \textbf{85.8}         & \textbf{84.1}         & \textbf{73.2}          & \textbf{86.8}          \\ \hline
\textit{Qwen2.5-7B}   & 94.1        & 91.7        & 90.8         & 89.1         & 80.7         & 63.6          & 85.0          \\
ThinK, 70\%           & 86.7        & 80.6        & 80.5         & 81.7         & 67.1         & 60.4          & 62.8          \\
\rowcolor{gray!20} LeanK, 70\%           & \textbf{93.6}        & \textbf{89.8}        & \textbf{90.6}         & \textbf{88.5}         & \textbf{78.0}         & \textbf{64.5}          & \textbf{84.2}          \\ \hline
\end{tabular}
}
 \caption{Performance of different methods and models on RULER. Detailed results are in Appendix \ref{sec:appendix2}. }
 \vspace{-8pt}
\label{tab:ruler_performance}
\end{table}

\subsection{Effectiveness of LeanK}

Table \ref{tab:longbench_rst}, Table \ref{tab:ruler_performance} and Table \ref{tab:gsm_infinite_rst} present performance results on three benchmarks, respectively. LeanK's effectiveness is demonstrated in four aspects:

\begin{figure*}[ht]
    \centering
    \includegraphics[width=1\linewidth]{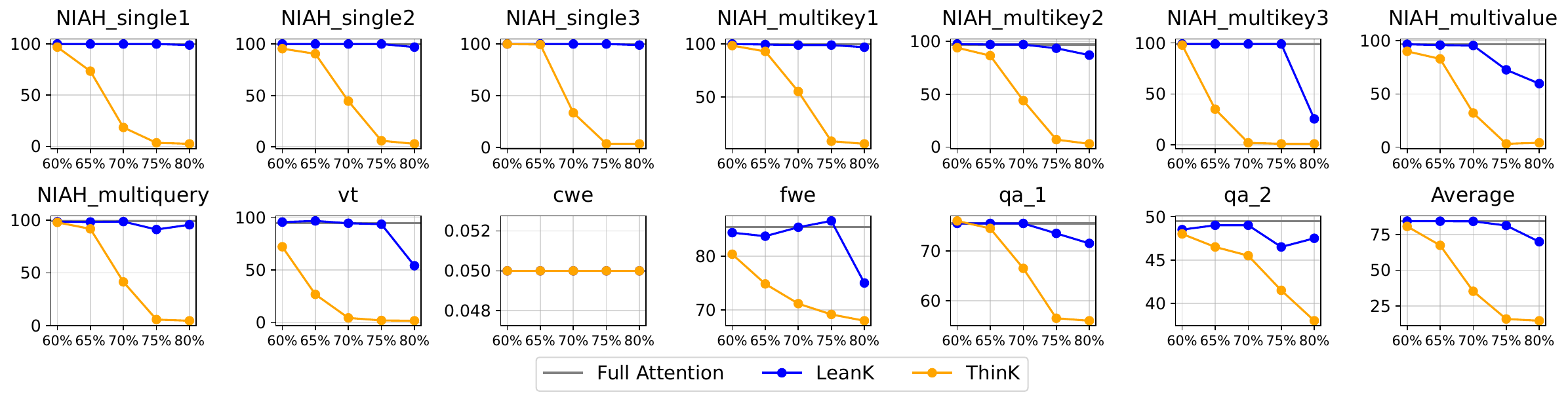}
    \vspace{-20pt}
    \caption{Comparison of performance on RULER 64K under different pruning ratios. }
    \vspace{-14pt}
    \label{fig:comparison}
\end{figure*}

\noindent \textbf{High Compression Ratio with Minimal Performance Loss.} 
Results show that LeanK maintains near-lossless performance under a 70\% compression ratio, while ThinK experiences significant degradation. On RULER, ThinK at a 70\% pruning ratio shows performance drops of 52.8\% on Llama and 26.1\% on Qwen; even at 60\% pruning, ThinK still suffers an 8.0\% drop on Llama. In contrast, LeanK only shows minimal drops of 0.3\% and 0.1\% on the two models under the same 70\% compression setting. On LongBench, ThinK drops by 5.7\% and 4.8\% under 70\% compression, while LeanK maintains strong performance with only 0.4\% and 3.1\% degradation.

\noindent \textbf{Strong Staticity in K Channels.} On RULER evaluation tasks, which features diverse input lengths ranging from 4K to 128K, LeanK demonstrates consistently strong performance using the learned static channel patterns. This indicates that the importance of K cache channels in pretrained LLMs exhibits a largely static nature.

\begin{table}[H]
    \centering
    \setlength{\tabcolsep}{1.8mm}
    \vspace{4pt}
    \resizebox{\columnwidth}{!}{
    \begin{tabular}{l|ccc|ccc|c}
        \toprule
          & \multicolumn{3}{c}{\textbf{Medium}} & \multicolumn{3}{c}{\textbf{Hard}} & \\ 
         \textbf{Methods} & \textbf{8K} & \textbf{16K} & \textbf{32K} & \textbf{8K} & \textbf{16K} & \textbf{32K} & \textbf{Avg.} \\
        \midrule
           \textit{Llama-3.1-8B} & 0.56 & 0.35 & 0.30 & 1.07 & 0.65 & 0.46 & 0.56 \\
         ThinK, 70\% & 0.26 & 0.17 & 0.15 & 0.28 & 0.14 & 0.12 & 0.19 \\
         \rowcolor{gray!20} LeanK, 70\% & \textbf{0.70} & \textbf{0.49} & \textbf{0.40} & \textbf{1.14} & \textbf{0.65} & \textbf{0.50} & \textbf{0.65} \\
        \midrule
          \textit{Qwen2.5-7B} & 1.08 & 0.92 & 1.06 & 1.01 & 0.93 & 0.88 & 0.98 \\
         ThinK, 70\% & 0.96 & 0.75 & 0.75 & 0.76 & 0.69 & 0.64 & 0.76 \\
         \rowcolor{gray!20} LeanK, 70\% & \textbf{1.06} & \textbf{0.86} & \textbf{0.76} & \textbf{1.03} & \textbf{0.85} & \textbf{0.74} & \textbf{0.88} \\
        \bottomrule  
    \end{tabular}}
    \caption{Performance of different methods and models on GSM-Infinite. For each subset, we calculate AUC score for op=2,4,6,8,10,12 with 256 samples for each op. Detailed results are in Appendix \ref{sec:appendix}.}
    \vspace{-14pt}
    \label{tab:gsm_infinite_rst}
\end{table}

\noindent \textbf{Strong Generalizability.} As illustrated in Table \ref{tab:gsm_infinite_rst}, LeanK's learned channel patterns generalize effectively to long-generation reasoning tasks, which are typically sensitive to compression \cite{li2025quantization}. Under a 70\% pruning ratio, ThinK suffers performance drops of 67\% on Llama and 20\% on Qwen. LeanK outperforms ThinK on both models, and even improves the performance of Llama by 13\%. Given the growing importance of reasoning tasks, LeanK offers a promising approach to enhancing efficiency of model inference.

\noindent \textbf{Resilience Under Extreme Sparsity.} Furthermore, we applied our method to Llama-3.1-8B-Instruct model across varying compression ratios and tested performance on RULER 64K tasks. As illustrated in Figure \ref{fig:comparison}, LeanK consistently surpasses ThinK under diverse pruning ratios, while also maintaining higher performance under aggressive pruning settings.

\begin{table*}[htbp]
\centering
\noindent
\begin{minipage}{0.3\textwidth}
\resizebox{1\columnwidth}{!}{
    \begin{tabular}{lc|c}
        \toprule
         Method & K Ratio & Acc \\
        \midrule
           Original & - & 84.38 \\
          DuoAttn & 50\% & 83.94 \\
         DuoAttn + LeanK & 80\% & 83.53 \\ 
        \bottomrule
    \end{tabular}}
\captionof{table}{LeanK on top of DuoAttn. Performance on RULER 64K.}
\label{tab:duo}
\end{minipage}
\hfill
\begin{minipage}{0.3\textwidth}
\centering
\resizebox{1\columnwidth}{!}{
    \begin{tabular}{lc|c}
        \toprule
         Method & K Ratio & Acc \\
        \midrule
           Original & - & 84.38 \\
          Quest & - &  72.41 \\
         Quest + LeanK & 70\% &  75.14 \\
        \bottomrule
    \end{tabular}}
\captionof{table}{LeanK on top of Quest. Performance on RULER 64K.}
\label{tab:quest}
\end{minipage}
\hfill
\begin{minipage}{0.3\textwidth}
\resizebox{1\columnwidth}{!}{
    \begin{tabular}{lc|c}
        \toprule
         Method & K Ratio & Acc \\
        \midrule
           Original & - & 86.03 \\
          KIVI & - &  84.67 \\
         KIVI + LeanK & 70\% &  84.16 \\
        \bottomrule
    \end{tabular}}
\captionof{table}{LeanK on top of KIVI. Performance on RULER 32K.}
\label{tab:kivi}
\end{minipage}
\vspace{-14pt}
\end{table*}

\subsection{Efficiency of LeanK}

\begin{figure}[H]
    \centering
    \includegraphics[width=0.74\linewidth]{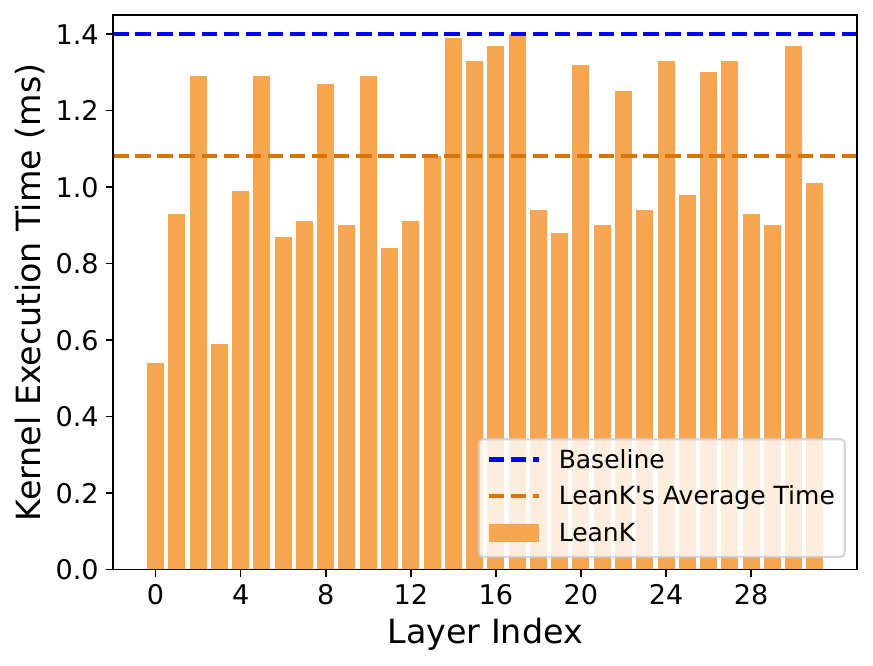}
    \vspace{-8pt}
    \caption{Kernel execution time of each layer on Llama-3.1-8B-Instruct. LeanK uses 70\% pruning ratio. Both Baseline and LeanK use Tilelang implementation. }
    \label{fig:kernel_time}
\end{figure}

\noindent \textbf{Kernel Design.} We implement custom decoding kernel to accelerate attention computation using TileLang~\cite{wang2025tilelang}. 
After loading the model weights, we group attention heads in each layer based on their remained channel count, and reorder the Q, K, V, O projection weights accordingly. For each group, we store a separate pruned K cache, along with the full K cache for sink and local tokens. 
At each decoding step, a fused decoding kernel is launched, which 
directly reads from the grouped K cache, performs FlashAttention, and outputs the final attention results. 
By reading significantly fewer K channels, the kernel reduces memory bandwidth usage and accelerates decoding. This approach achieves 1.3× and 1.6× average speedup in attention computation on Llama-3.1-8B-Instruct and Qwen2.5-7B-Instruct, respectively, as shown in Figure~\ref{fig:kernel_time} and Figure~\ref{fig:qwen_kernel_time}.

\noindent \textbf{GPU Memory Reduction.} Under a 70\% pruning ratio, LeanK achieves approximately 70\% GPU memory reduction in the K cache for long-context inputs. Moreover, when all channels of a head's K cache are pruned, the corresponding V cache can also be safely removed (\S \ref{sec:deployment}). This occurs in approximately 18\% heads in Llama-3.1-8B-Instruct and 16\% in Qwen2.5-7B-Instruct. We evaluate the memory reduction on a single 80GB A100 GPU with an input length of 4096 and an output length of 1024. As shown in Figure~\ref{fig:memory}, LeanK enables a 20\% larger batch size and saves approximately 10GB of memory when the batch size is 64.

\noindent \textbf{Increased End-to-end Throughput. } 
Combining our efficient decoding kernel, enlarged batch size and a more efficient KV cache management strategy, LeanK achieves a 1.2× increase in end-to-end throughput on Llama-3.1-8B-Instruct, as shown in Table~\ref{tab:throughput} \footnote{Huggingface Baseline uses a batch size of 52 since larger batch sizes would lead to OOM. }. 

\begin{table}[H]
    \centering
    \vspace{4pt}
    \resizebox{1\columnwidth}{!}{
    \begin{tabular}{l|ccc|c}
    \toprule
         & Batch size & Gen length & Gen Time & Throughput \\ 
         \midrule
         Baseline & 52 & 128 & 47.27 s & 141 tokens/s \\
        LeanK & 64 & 128 & 47.62 s & 172 tokens/s \\
        \bottomrule
    \end{tabular}
    }
    \vspace{-8pt}
    \caption{End-to-end generation time and throughput. Tested with Huggingface transformers framework, with input sequence length 4096. }
    \label{tab:throughput}
    \vspace{-8pt}
\end{table}

\subsection{Orthogonality with Other Methods}
\label{sec:orthogonal}

We emphasize that K channel sparsity is orthogonal to existing approaches. LeanK can be combined with other KV cache optimization methods for further acceleration, especially in resource-constrained environments. 

\noindent \textbf{DuoAttention}. DuoAttention~\cite{xiao2024duoattentionefficientlongcontextllm} is a KV cache eviction method that categorizes heads into Streaming heads and Retrieval heads, evicting the KV cache of the former. LeanK can be applied to the remaining Retrieval Heads, boosting KV cache memory reduction from 50\% to 65\%, without performance degradation (Table \ref{tab:duo}).

\noindent \textbf{Quest}: Quest~\cite{tang2024questqueryawaresparsityefficient} is a KV cache selective reading method that identifies and loads only critical pages during decoding. LeanK can be applied to both the critical page selection and loading phases, reducing memory reads by 70\% and improving model accuracy (Table~\ref{tab:quest}).

\noindent \textbf{KIVI}: KIVI~\cite{liu2024kivi} is a KV cache quantization method. LeanK can be applied beforehand to prune unimportant KV entries, and then KIVI quantizes remaining cache. This combination improves compression ratio from 5.3× to 9.7×, using 2-bit quantization for both K and V (Table~\ref{tab:kivi})\footnote{When our method is not applied and only KIVI is used, we encounter an OOM issue with the official implementation at 64K input length, so we conduct experiments at 32K instead.}.

\subsection{Ablation Study}
\label{sec:ablation_studies}

LeanK involves two design decisions for pruning K cache channels: (1) employing a learned pruning mask rather than relying on norm-based selection, and (2) allocating pruning budgets globally across all heads, resulting in varying numbers of retained channels per head. In contrast, ThinK utilizes a uniform budget across heads and a dynamic, norm-based mask. To clearly assess these design choices, we apply LeanK's per-head learned budget but replace its learned mask with the dynamic norm-based mask, with results presented in Table~\ref{tab:alloc_budget_main}. We observe the per-head budget alone significantly boosts accuracy from 35.29 to 76.59, underscoring the advantage of adaptive budget allocation across heads. Furthermore, replacing the dynamic norm-based mask with LeanK’s learned mask further boosts accuracy to 84.10, demonstrating the effectiveness of the learned sparsity. 

\begin{table}[h]
    \centering
    \setlength{\tabcolsep}{1.8mm}
    \resizebox{1\columnwidth}{!}{
    \begin{tabular}{l|c|c}
        \toprule
        Method & Ratio & Acc \\
        \midrule
        Original & - & 84.38 \\
        Uniform Budget, Dynamic Mask (ThinK) & 70\% & 35.29 \\
        Per-head Budget, Dynamic Mask & 70\% & 76.59 \\
        Per-head Budget, Learned Mask (LeanK) & 70\% & 84.10 \\
        \bottomrule
    \end{tabular}
    }
    \vspace{-8pt}
    \caption{Evaluation results on Llama-3.1-8B-Instruct, with accuracy measured on RULER 64K.}
    \vspace{-16pt}
    \label{tab:alloc_budget_main}
\end{table}

%% file: sections/analysis.tex
\section{Analysis}

\begin{figure*}[ht]
    \centering
    \begin{minipage}[t]{0.3\textwidth}
    \vspace{-\baselineskip}
        \includegraphics[width=\textwidth]{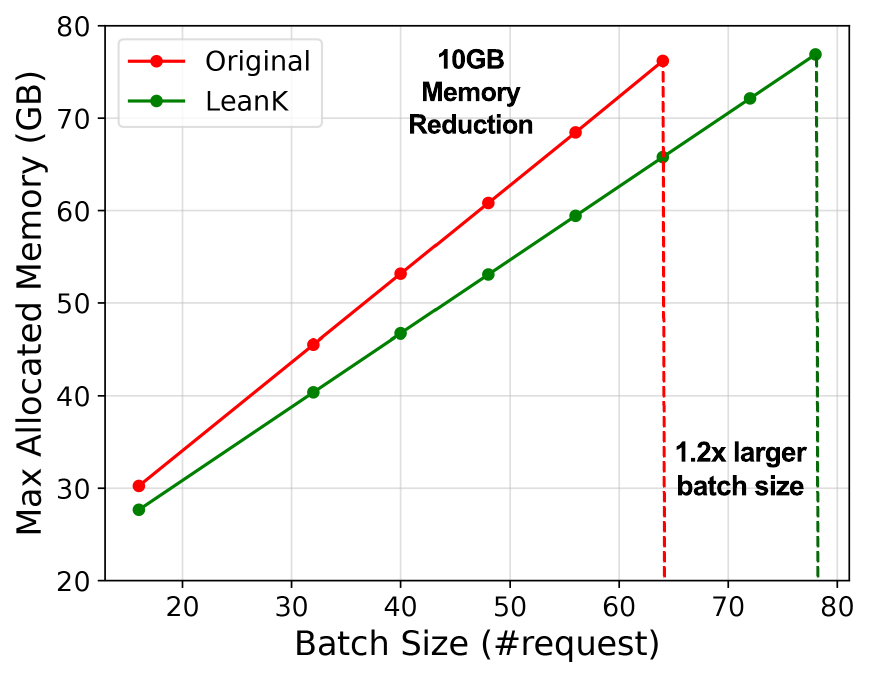} 
        \caption{Batch Size and Memory. LeanK enables a 20\% larger batch size, saving 10GB memory. }
        \label{fig:memory}
    \end{minipage}\hfill
    \begin{minipage}[t]{0.318\textwidth}
    \vspace{-\baselineskip}
        \includegraphics[width=\textwidth]{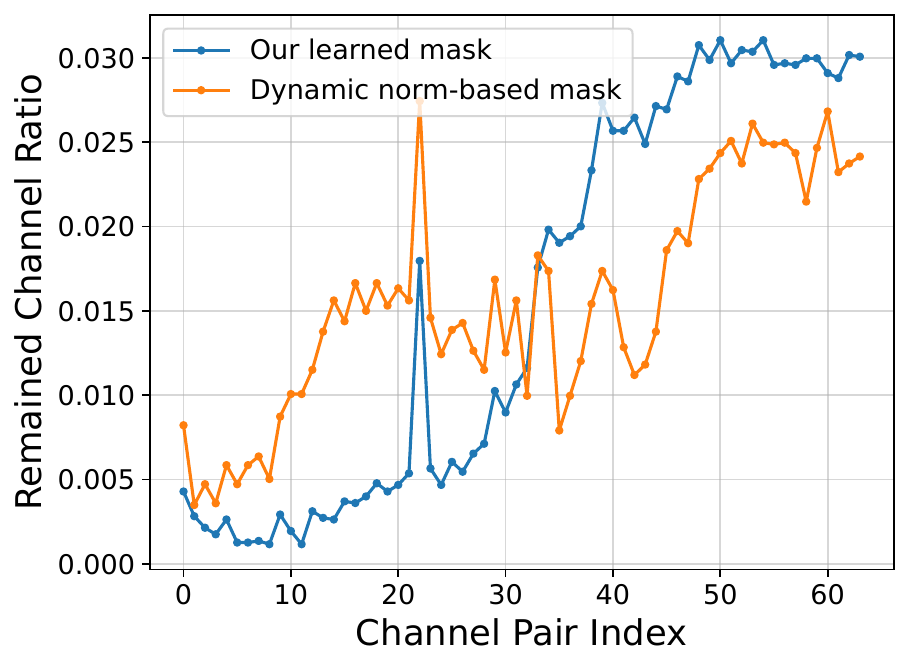} 
        \caption{Channel Pair Index and Remained Channel Ratio on Llama-3.1-8B-Instruct. }
        \label{fig:freqency}
    \end{minipage}\hfill
    \begin{minipage}[t]{0.315\textwidth}
    \vspace{-\baselineskip}
        \includegraphics[width=\textwidth]{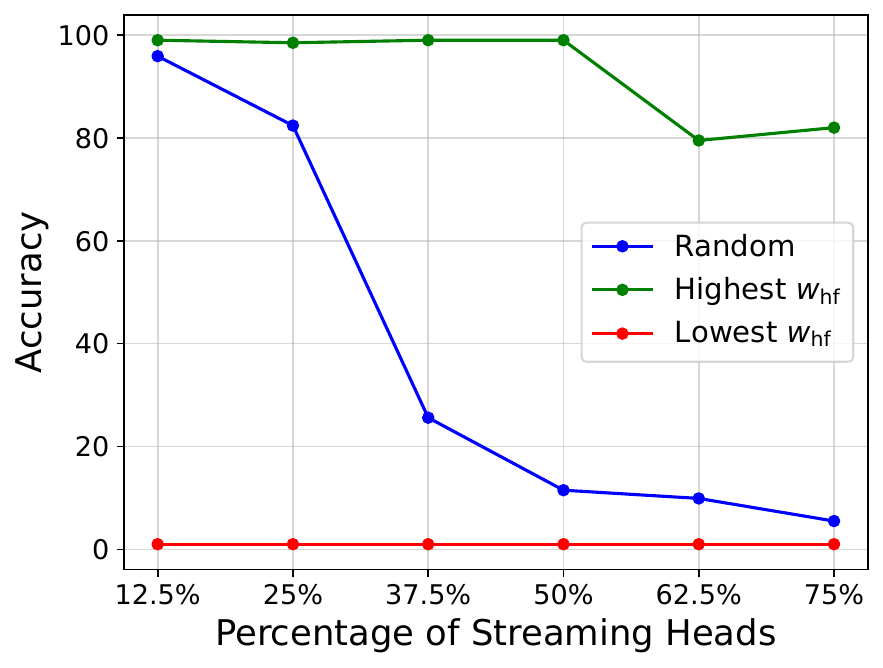}
        \caption{Converting heads with highest or lowest $w_{\text{hf}}$ values into streaming heads and performance.}
        \label{fig:classify_high_freq}
    \end{minipage} 
    \vspace{-14pt}
\end{figure*}

In this section, we analyze the learned channel importance distribution to gain deeper insight into model's behavior.

\subsection{Frequency and Channel Importance}
\label{sec:rope}

RoPE assigns specific frequencies to every pair of channels in the K matrix. We introduce the term "channel pair index" to indicate each pair of channels, where smaller indices correspond to higher frequencies and larger indices to lower frequencies. Both Llama-3.1-8B-Instruct and Qwen2.5-7B-Instruct contain 64 channel pairs per head. Figures \ref{fig:freqency} and \ref{fig:freqency-qwen} illustrate the retained ratio of channels relative to their channel pair indices.

We have two observations: (1) Channel pairs with lower frequencies generally exhibit higher importance, aligning with previous research indicating that semantic information crucial for long-context understanding is encoded mainly in low-frequency channels \cite{barbero2025roundroundgomakes}. Our pruning method retains significantly fewer high-frequency channels than dynamic norm-based methods 
. (2) However, exceptions exist -- channel pair 22 in Llama and channel pair 31 in Qwen, despite being high-frequency, show considerable importance. 
A further investigation into their specific functions remains for future work.

\subsection{High Frequency Ratio of Each Head}
\label{sec:high-frequency-sensitivity}

Inspired by Retrieval Heads identification methods such as \citet{wu2024retrieval} and the RoPE frequency analysis in Section~\ref{sec:rope}, we want to investigate the relationship between {retrieval ability} of different heads and the {ratio of their high frequency components} $w_{\text{hf}}$, defined as:
\begin{equation*}
w_{\text{hf}}=\frac{||\bm{q}_{[\text{:high}]}\bm{K}_{[\text{:high}]}^T||_2}{||\bm{q}\bm{K}^T||_2}
\end{equation*}

where $\bm{q}_{[\text{:high}]}$ and $\bm{K}_{[\text{:high}]}$ represent the higher-frequency half of channel dimensions.\footnote{We use half (high=32) as the boundary between high and low frequency components for simplicity. } 
Since high frequency channels mainly contain less informative noises, heads with higher $w_{\text{hf}}$ are more likely to be influenced by such noises and might be less crucial for capturing semantic information. 
We compute $w_{\text{hf}}$ using a single input sequence from RULER NIAH\_multikey3 task for Llama-3.1-8B-Instruct, and examine the importance of heads with different $w_{\text{hf}}$ values through head pruning. 

We convert heads with highest or lowest $w_{\text{hf}}$ into streaming heads and evaluate performance on NIAH\_multikey3 task. For comparison, we randomly select the same number of heads to convert and report average performance over four trials. Results in Figure \ref{fig:classify_high_freq} suggest that {heads with low \(w_{hf}\) are crucial for long-context understanding} while {heads with high \(w_{hf}\) could be pruned with minimal impact}. 
Results in Appendix \ref{sec:head_classification_details} show that it also generalizes well on other tasks. This opens the opportunity of a effective, training-free head pruning strategy with minimal calibration cost.

%% file: sections/conclusion.tex
\section{Conclusion}

We propose LeanK, a learning-based method for pruning the channel dimension of K cache to enable efficient LLM decoding. LeanK employs a double-stage optimization process to learn a static pruning mask. Experiments demonstrate that LeanK reduces GPU memory usage by up to 70\% for the K cache and 16\%--18\% for the V cache, achieving a 1.45$\times$ speedup during inference, while preserving model accuracy.

%% file: sections/appendix.tex
\clearpage
\appendix

\section{Quantifying K Channel Staticity}
\label{appendix:k-channel-staticity}

We aim to quantify the \textit{staticity} of K channels across various tasks and input lengths. For a specific channel $i$ within an attention head, given an input sample, we measure the following norm ratio:

\begin{equation*}
r_{i} = \frac{||\bm{Q}_{[i]}\bm{K}_{[i]}^T||_2}{||\bm{Q}\bm{K}^T||_2},
\end{equation*}
where $\bm{Q}$ represents the query matrix corresponding to an observation window located at the last part of the input, $\bm{K}$ is the complete key matrix, and $\bm{Q}_{[i]}$, $\bm{K}_{[i]}$ denote the $i$-th channel of $\bm{Q}$ and $\bm{K}$, respectively. The ratio $r_i$ captures the relative importance of channel $i$ within an attention head.

We aggregate these channel-wise ratios $r_i$ into a single vector $\bm{r}$ of shape $(L \times n \times d,)$, where $L$, $n$, and $d$ correspond to the number of layers, attention heads per layer, and channels per head, respectively.

To evaluate staticity across tasks, we measure $\bm{r}$ separately for different tasks and compute the Pearson correlation coefficient between these vectors. Specifically, for tasks 1 and 2, we obtain vectors $\bm{r}^{(1)}$ and $\bm{r}^{(2)}$. If channel importance significantly differs between tasks, we would expect a low correlation between $\bm{r}^{(1)}$ and $\bm{r}^{(2)}$. However, as illustrated in Figure \ref{fig:motivation}, Pearson correlations between different RULER tasks consistently remain close to 1. This indicates that the high-importance channels in one task generally remain highly important in another, highlighting a static sparsity pattern in the K channels. Similar results are observed across varying input lengths, reinforcing the notion of channel staticity.

\section{Comparison with Double Sparsity}
\label{sec:ds_comparison}

Double Sparsity~\cite{yang2024posttrainingsparseattentiondouble} proposes a KV selective-reading strategy based on offline identification of outlier channel dimensions. Specifically, it computes the norm of the QK product to select high-norm (outlier) channels, which are then used during inference to retrieve critical tokens.

In this section, we compare our learned channel importance scores with the outlier channel selection criterion used in Double Sparsity to evaluate whether their method can be used for channel pruning. As shown in Table~\ref{tab:ds_rst_rough}, we find that Double Sparsity's method exhibits several limitations that lead to suboptimal performance:

\begin{enumerate} 
\item The outlier channel selection process is not inherently designed for channel-wise pruning, and may lack careful design considerations.
\item During offline norm collection, the method splits QK product on context dimension into chunks of smaller sizes, which overlooks the unstructured composition of QK channel dimension (namely, the attention sink and local window) and may lead to inaccurate norm calculation. 
\item As discussed in Section~\ref{sec:motiv}, relying solely on norm (magnitude) is an insufficient proxy for estimating channel importance. This limitation may restrict the pruning ratio of local norm-based pruning methods.
\end{enumerate}

\begin{table}[H]
    \centering
    \setlength{\tabcolsep}{1.8mm}
    \vspace{-8pt}
    \resizebox{0.8\columnwidth}{!}{
    \begin{tabular}{c|lc|c}
        \toprule
        & Method & Ratio & Acc \\
        \midrule
          & Original & - & 84.4 \\
        Llama-3.1-8B-Instruct & DS & 60\% & 29.9 \\
        & DS & 70\% & 14.0 \\
        & LeanK & 70\% & 84.1 \\
        \bottomrule
    \end{tabular}
    }
    \caption{Comparison with Double Sparsity. Methods are tested on RULER 64K, with 200 samples taken from each subtask. Detailed results are in Table~\ref{tab:ds_rst}. }
    \vspace{-16pt}
    \label{tab:ds_rst_rough}
\end{table}

\begin{table*}[t]
    \centering
    \setlength{\tabcolsep}{1.8mm}
    \resizebox{2\columnwidth}{!}{
    \begin{tabular}{lc|ccccccccccccc|c}
        \toprule
        & & \multicolumn{14}{c}{\textbf{Llama-3.1-8B-Instruct}} \\ 
        Method & \textbf{Ratio} & \textbf{niah\_s1} & \textbf{niah\_s2} & \textbf{niah\_s3} & \textbf{niah\_mk1} & \textbf{niah\_mk2} & \textbf{niah\_mk3} & \textbf{niah\_mv} & \textbf{niah\_mq} & \textbf{vt} & \textbf{cwe} & \textbf{fwe} & \textbf{qa\_1} & \textbf{qa\_2} & Avg. \\
        \midrule
          Original & - &  100.0 & 100.0 & 100.0 & 100.0 & 97.0 & 99.0 & 97.0 & 99.1 & 94.5 & 0.1 & 85.3 & 75.5 & 49.5 & 84.4 \\
        Highest $w_{\text{hf}}$ & 25\% & 99.5 & 99.5 & 100.0 & 100.0 & 96.5 & 98.5 & 96.6 & 98.6 & 93.5 & 0.15 & 85.5 & 76.5 & 49.5 & 84.2 \\
         \rowcolor{gray!20} Highest $w_{\text{hf}}$ & 50\% & 99.5 & 100.0 & 100.0 & 100.0 & 96.0 & 99.0 & 93.8 & 98.9 & 92.2 & 0.4 & 83.0 & 75.5 & 48.5 & 83.6 \\
          Lowest $w_{\text{hf}}$ & 25\% & 2.5 & 3.0 & 3.5 & 5.5 & 3.0 & 1.0 & 4.5 & 5.0 & 1.5 & 0.1 & 67.2 & 53.0 & 40.0 & 14.6 \\
          Random & 25\% & 99.6 & 98.2 & 99.2 & 99.5 & 93.6 & 82.4 & 71.7 & 93.8 & 70.1 & 0.1 & 80.1 & 73.9 & 46.8 & 77.6 \\
         \rowcolor{gray!20} DuoAttn & 50\% & 100.0 & 100.0 & 100.0 & 99.0 & 96.5 & 99.0 & 95.8 & 99.3 & 91.2 & 0.05 & 84.5 & 76.0 & 50.0 & 84.0 \\
        \bottomrule
    \end{tabular}}
    \caption{Head pruning and classification based on \textbf{high frequency ratio} of each head. }
    \label{tab:hfs}
\end{table*}

\section{Channel Frequency Analysis for Qwen2.5-7B-Instruct}
\label{sec:qwen_analysis}

\begin{figure}[H]
    \centering
    \includegraphics[width=0.8\linewidth]{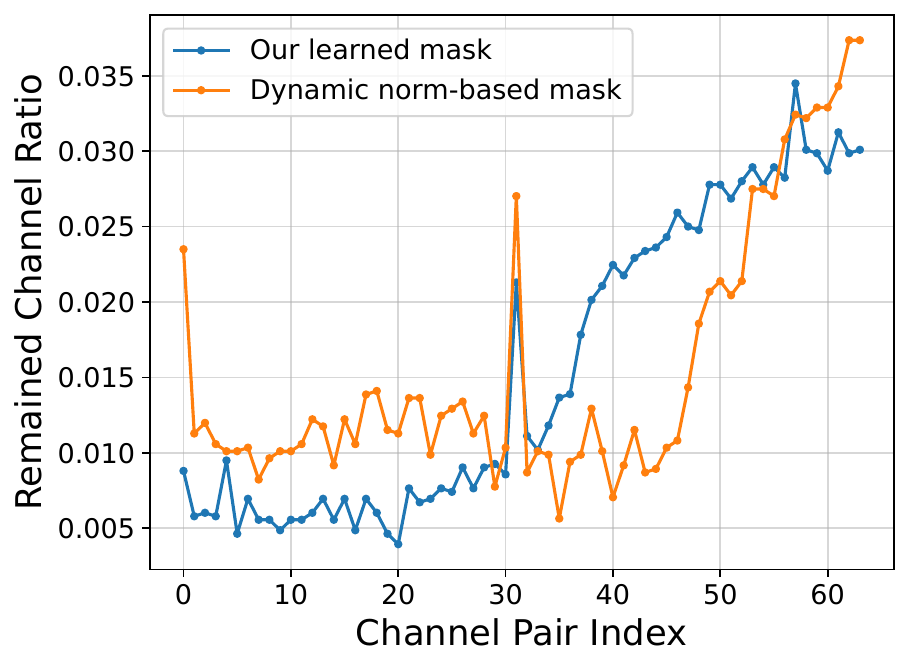} 
    \caption{Remained Ratio and Channel Pair Index on Qwen2.5-7B-Instruct.}
    \label{fig:freqency-qwen}
\end{figure}

We visualize the learned mask for Qwen2.5-7B-Instruct in Figure \ref{fig:freqency-qwen}. Similar as Section \ref{sec:rope}, we observed that LeanK's learned pruning pattern preserves more low frequency channels compared with ThinK's norm-based method, which might contribute to LeanK's effectiveness. 

Furthermore, a similar outlier channel with channel pair index 31 appears to have relatively high norm and large impact on model performance. 

\section{Head Pruning based on High Frequency Ratio}
\label{sec:head_classification_details}

In Section~\ref{sec:high-frequency-sensitivity}, we define the \textit{High Frequency Ratio} of each attention head as:
\begin{equation*}
w_{\text{hf}} = \frac{||\bm{q}_{[\text{:high}]} \bm{K}_{[\text{:high}]}^T||_2}{||\bm{q} \bm{K}^T||_2}
\end{equation*}

We compute $w_{\text{hf}}$ using a single 64K-token input from the RULER VT task. We observe that this score effectively distinguishes retrieval heads. To evaluate its utility, we modify Llama-3.1-8B-Instruct by converting a subset of heads into streaming heads based on their $w_{\text{hf}}$ values. Specifically, we convert heads with the highest or lowest $w_{\text{hf}}$ values and measure end-to-end performance on RULER 64K. For comparison, we also convert an equal number of randomly selected heads into streaming heads, repeating the experiment four times and reporting the average result (Table~\ref{tab:hfs}).

Our findings show that converting heads with the lowest $w_{\text{hf}}$ severely degrades performance, while converting the top 50\% highest $w_{\text{hf}}$ heads has minimal impact. This matches the performance of DuoAttention~\cite{xiao2024duoattentionefficientlongcontextllm}, which requires an additional training phase. In contrast, our method achieves comparable results using only a single input sequence for calibration.

\section{Necessity of Training Stage 2}
\label{sec:stage2_detail}

We verified that applying Top-K on scaling factor \(\alpha\) trained from the training stage 1 leads to sub-optimal result on model performance, results are shown in Table \ref{tab:2nd_stage_rough}, suggesting that the second training stage is necessary for pruning under predefined sparsity ratios. 

The primary reason for the performance disparity lies in the misalignment between the scaling operation in Stage1 and the ultimate objective of channel masking for deployment. For instance, some channels may be robust to scaling, but entirely masking them out can lead to severe performance degradation. Stage2 could help avoid these channels from being pruned. 

Furthermore, Direct Top-K on \(\alpha\) violates alignment requirements (aligning to multiplies of 16 or 32) and might be inefficient for hardware execution. 

\begin{table}[H]
    \centering
    \setlength{\tabcolsep}{1.8mm}
    \resizebox{1\columnwidth}{!}{
    \begin{tabular}{c|lc|c}
        \toprule
        & Method & Ratio & Acc \\
        \midrule
          & Original & - & 80.7 \\
        Qwen2.5-7B-Instruct & w/o 2nd stage & 70\% & 70.7 \\
        & w/ 2nd stage & 70\% & 78.0 \\
        \bottomrule
    \end{tabular}
    }
    \caption{Necessity of 2nd stage of training. Methods are tested on RULER 64K, with 200 samples taken from each subtask. Using only the first stage with a 70\% pruning ratio yields a RULER 64K accuracy of just 70.73. With the second stage added, accuracy improves to 77.97 on Qwen2.5-7B-Instruct. Detailed results are in Table~\ref{tab:2ndstage_details}. }
    \label{tab:2nd_stage_rough}
\end{table}

\section{Kernel Benchmarking on Qwen}
\label{sec:qwen_kernel_time}

\begin{figure}[H]
    \centering
    \includegraphics[width=0.8\linewidth]{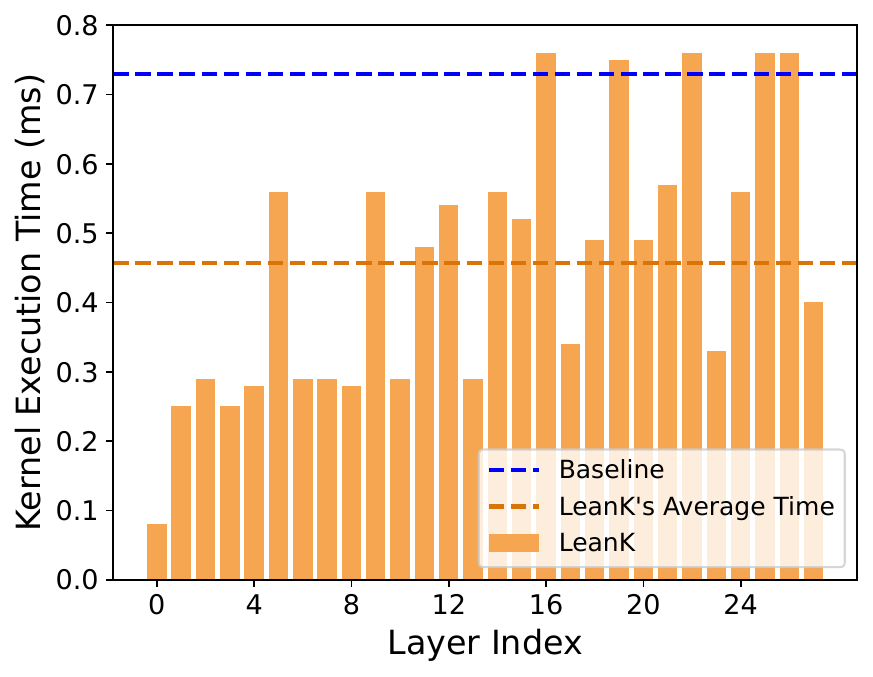}
    \caption{Kernel execution time of each layer on Qwen2.5-7B-Instruct. LeanK uses 70\% pruning ratio. }
    \label{fig:qwen_kernel_time}
\end{figure}

LeanK could achieve a 1.6x speedup on attention computation on Qwen2.5-7B-Instruct. Execution time of each layer is shown in Figure~\ref{fig:qwen_kernel_time}.

\section{Full Evaluation Results on RULER}
\label{sec:appendix2}

Full evaluation results on RULER \cite{hsieh2024rulerwhatsrealcontext} are shown in Table \ref{tab:ruler_full2}. We evaluate all tasks in RULER across input lengths ranging from 4K to 128K. For Qwen testing on input lengths larger than 32K, we apply Yarn extrapolation with a factor of 4 as suggested by its official documentation. 

\begin{table*}[t]
    \centering
    \setlength{\tabcolsep}{1.8mm}
    \resizebox{2\columnwidth}{!}{
    \begin{tabular}{c|lcccccccccccccc|c}
        \toprule
        & & & \multicolumn{13}{c}{\textbf{Llama-3.1-8B-Instruct}} \\ 
        & Method & \textbf{Ratio} & \textbf{niah\_s1} & \textbf{niah\_s2} & \textbf{niah\_s3} & \textbf{niah\_mk1} & \textbf{niah\_mk2} & \textbf{niah\_mk3} & \textbf{niah\_mv} & \textbf{niah\_mq} & \textbf{vt} & \textbf{cwe} & \textbf{fwe} & \textbf{qa\_1} & \textbf{qa\_2} & Avg. \\
        \midrule
          & Original & - & 100.0 & 100.0 & 100.0 & 99.5 & 99.0 & 100.0 & 99.8 & 99.6 & 99.4 & 99.6 & 93.7 & 85.0 & 61.0 & 95.1 \\
        4K & ThinK & 60\% &  93.0 & 88.0 & 99.5 & 88.5 & 97.0 & \textbf{99.0} & 90.0 & 92.0 & 76.9 & 99.1 & 93.3 & \textbf{86.5} & 61.0 & 89.5 \\
        & ThinK & 70\% & 31.0 & 49.5 & 43.0 & 66.0 & 80.0 & 35.5 & 49.3 & 53.9 & 23.6 & 97.4 & 84.5 & \textbf{86.5} & 58.0 & 58.3 \\
        & Ours & 70\% & \textbf{100.0} & \textbf{100.0} & \textbf{100.0} & \textbf{99.5} & \textbf{99.5} & \textbf{99.0} & \textbf{99.8} & \textbf{99.6} & \textbf{99.3} & \textbf{99.6} & \textbf{94.3} & \textbf{86.5} & \textbf{61.5} & \textbf{95.3} \\
        \midrule
          & Original & - & 100.0 & 100.0 & 100.0 & 100.0 & 99.5 & 99.5 & 99.5 & 99.8 & 98.9 & 94.4 & 84.5 & 78.0 & 56.0 & 93.1 \\
        8K & ThinK & 60\% & 75.5 & 87.5 & 98.0 & 87.5 & 99.0 & 98.5 & 87.1 & 93.0 & 76.2 & 42.3 & 84.5 & 76.0 & \textbf{56.5} & 81.7 \\
        & ThinK & 70\% & 12.5 & 33.0 & 19.5 & 64.0 & 66.0 & 12.5 & 29.3 & 46.9 & 15.5 & 20.4 & 68.7 & 68.5 & 53.0 & 39.2 \\
        & Ours & 70\% & \textbf{100.0} & \textbf{100.0} & \textbf{100.0} & \textbf{100.0} & \textbf{100.0} & \textbf{99.5} & \textbf{99.3} & \textbf{99.6} & \textbf{99.2} & \textbf{94.7} & \textbf{87.0} & \textbf{78.5} & 56.0 & \textbf{93.4} \\
        \midrule
          & Original & - & 100.0 & 100.0 & 100.0 & 99.5 & 100.0 & 99.5 & 99.4 & 98.9 & 99.3 & 53.3 & 90.7 & 79.5 & 53.0 & 90.2 \\
        16K & ThinK & 60\% & 75.0 & 88.5 & 99.0 & 95.0 & 99.5 & 98.5 & 92.4 & 94.8 & 66.8 & 0.9 & \textbf{96.7} & 78.0 & 51.5 & 79.7 \\
        & ThinK & 70\% & 9.5 & 28.5 & 23.0 & 60.5 & 65.5 & 5.0 & 31.5 & 46.8 & 9.6 & 0.4 & 92.3 & 66.5 & 48.5 & 37.1 \\
        & Ours & 70\% & \textbf{100.0} & \textbf{100.0} & \textbf{100.0} & \textbf{99.5} & \textbf{100.0} & \textbf{99.5} & \textbf{99.0} & \textbf{98.1} & \textbf{98.6} & \textbf{36.0} & 91.7 & \textbf{79.0} & \textbf{53.0} & \textbf{88.8} \\
        \midrule
          & Original & - & 100.0 & 100.0 & 100.0 & 100.0 & 99.0 & 100.0 & 98.9 & 99.4 & 97.6 & 2.7 & 93.3 & 76.0 & 51.5 & 86.0 \\
        32K & ThinK & 60\% & 88.0 & 94.0 & \textbf{100.0} & 97.5 & 98.5 & \textbf{100.0} & 95.3 & 98.5 & 70.0 & 0.0 & \textbf{96.2} & 75.5 & 49.5 & 81.8 \\
        & ThinK & 70\% & 9.5 & 32.5 & 21.0 & 58.5 & 54.5 & 2.0 & 37.9 & 48.6 & 9.2 & 0.0 & 91.7 & 62.5 & 45.0 & 36.4 \\
        & Ours & 70\% & \textbf{100.0} & \textbf{100.0} & \textbf{100.0} & \textbf{99.5} & \textbf{99.5} & \textbf{100.0} & \textbf{98.4} & \textbf{98.5} & \textbf{97.0} & \textbf{1.2} & 94.2 & \textbf{76.5} & \textbf{51.0} & \textbf{85.8} \\
        \midrule
          & Original & - & 100.0 & 100.0 & 100.0 & 100.0 & 97.0 & 99.0 & 97.0 & 99.1 & 94.5 & 0.1 & 85.3 & 75.5 & 49.5 & 84.4 \\
        64K & ThinK & 60\% & 97.0 & 95.5 & \textbf{100.0} & 98.5 & 94.0 & 98.0 & 90.1 & 97.8 & 72.0 & 0.1 & 80.3 & \textbf{76.0} & 48.0 & 80.6 \\
        & ThinK & 70\% & 18.5 & 44.5 & 33.5 & 55.0 & 44.0 & 2.0 & 32.1 & 41.4 & 4.5 & \textbf{0.1} & 71.2 & 66.5 & 45.5 & 35.3 \\
        & Ours & 70\% & \textbf{100.0} & \textbf{100.0} & \textbf{100.0} & \textbf{99.0} & \textbf{97.0} & \textbf{99.0} & \textbf{95.6} & \textbf{98.5} & \textbf{94.3} & \textbf{0.1} & \textbf{85.3} & 75.5 & \textbf{49.0} & \textbf{84.1} \\
        \midrule
          & Original & - & 100.0 & 99.0 & 100.0 & 97.0 & 75.0 & 53.5 & 93.1 & 97.3 & 54.5 & 0.2 & 76.3 & 71.5 & 37.5 & 73.5 \\
        128K & ThinK & 60\% & 98.5 & 97.0 & 99.5 & \textbf{97.0} & 71.0 & 26.0 & \textbf{91.9} & \textbf{97.4} & 40.0 & 0.3 & \textbf{80.2} & 70.0 & \textbf{38.5} & 69.8 \\
        & ThinK & 70\% & 29.0 & 67.0 & 28.9 & 75.5 & 40.0 & 0.0 & 50.9 & 52.9 & 8.3 & 0.5 & 73.2 & 59.0 & 35.0 & 40.0 \\
        & Ours & 70\% & \textbf{100.0} & \textbf{99.0} & \textbf{100.0} & 96.5 & \textbf{75.0} & \textbf{51.5} & 88.1 & 96.6 & \textbf{61.6} & \textbf{0.4} & 74.8 & \textbf{70.5} & 38.0 & \textbf{73.2} \\
    \end{tabular}}

    \resizebox{2\columnwidth}{!}{
    \begin{tabular}{c|lcccccccccccccc|c}
        \toprule
        & & & \multicolumn{13}{c}{\textbf{Qwen2.5-7B-Instruct}} \\ 
        & Method & \textbf{Ratio} & \textbf{niah\_s1} & \textbf{niah\_s2} & \textbf{niah\_s3} & \textbf{niah\_mk1} & \textbf{niah\_mk2} & \textbf{niah\_mk3} & \textbf{niah\_mv} & \textbf{niah\_mq} & \textbf{vt} & \textbf{cwe} & \textbf{fwe} & \textbf{qa\_1} & \textbf{qa\_2} & Avg. \\
        \midrule
          & Original & - & 100.0 & 100.0 & 100.0 & 100.0 & 100.0 & 100.0 & 94.6 & 100.0 & 100.0 & 99.7 & 84.2 & 84.5 & 60.0 & 94.1 \\
        4K & ThinK & 70\% & 97.0 & 94.0 & 98.0 & 96.0 & 95.0 & 78.0 & 82.4 & 94.4 & 80.5 & 98.6 & 73.3 & \textbf{84.0} & 55.5 & 86.7 \\
        & Ours & 70\% & \textbf{100.0} & \textbf{100.0} & \textbf{100.0} & \textbf{100.0} & \textbf{100.0} & \textbf{100.0} & \textbf{93.8} & \textbf{100.0} & \textbf{99.7} & \textbf{99.4} & \textbf{80.0} & \textbf{84.0} & \textbf{60.5} & \textbf{93.6} \\
        \midrule
          & Original & - & 100.0 & 100.0 & 100.0 & 100.0 & 100.0 & 99.0 & 89.8 & 99.9 & 100.0 & 94.0 & 88.0 & 69.5 & 52.0 & 91.7 \\
        8K & ThinK & 70\% & 98.5 & 90.0 & 98.0 & 93.0 & 93.5 & 60.5 & 72.1 & 90.5 & 69.3 & 86.3 & 83.7 & 61.5 & 50.5 & 80.6 \\
        & Ours & 70\% & \textbf{100.0} & \textbf{100.0} & \textbf{100.0} & \textbf{100.0} & \textbf{100.0} & \textbf{98.0} & \textbf{88.0} & \textbf{99.6} & \textbf{99.1} & \textbf{90.6} & \textbf{89.5} & \textbf{67.5} & \textbf{53.0} & \textbf{89.8} \\
        \midrule
          & Original & - & 100.0 & 100.0 & 100.0 & 99.5 & 100.0 & 92.5 & 95.1 & 99.9 & 98.1 & 85.4 & 92.0 & 63.5 & 54.0 & 90.8 \\
        16K & ThinK & 70\% & 97.5 & 98.0 & 97.5 & 93.0 & 94.0 & 59.5 & 77.9 & 89.4 & 69.1 & 66.0 & 91.7 & 59.5 & 53.5 &  80.5 \\
        & Ours & 70\% & \textbf{100.0} & \textbf{100.0} & \textbf{100.0} & \textbf{99.5} & \textbf{99.5} & \textbf{93.0} & \textbf{92.9} & \textbf{99.8} & \textbf{96.4} & \textbf{84.1} & \textbf{94.7} & \textbf{64.0} & \textbf{54.0} & \textbf{90.6} \\
        \midrule
          & Original & - & 100.0 & 100.0 & 99.5 & 99.0 & 98.5 & 94.0 & 91.4 & 98.9 & 96.9 & 67.7 & 87.8 & 68.0 & 57.0 & 89.1 \\
        32K & ThinK & 70\% & 99.5 & 99.0 & 98.0 & 98.0 & 94.0 & 67.5 & 70.9 & 91.4 & 75.7 & 57.1 & \textbf{91.0} & \textbf{66.0} & 53.5 & 81.7 \\
        & Ours & 70\% & \textbf{100.0} & \textbf{100.0} & \textbf{100.0} & \textbf{99.0} & \textbf{97.5} & \textbf{93.5} & \textbf{85.3} & \textbf{98.9} & \textbf{96.0} & \textbf{67.8} & 90.5 & 65.5 & \textbf{56.0} & \textbf{88.5} \\
        \midrule
          & Original & - & 100.0 & 98.0 & 98.0 & 95.5 & 82.5 & 47.5 & 82.8 & 97.4 & 95.3 & 55.4 & 82.5 & 69.5 & 44.5 & 80.7 \\
        64K & ThinK & 70\% & 95.0 & 91.5 & 95.0 & 85.0 & 63.0 & 24.5 & 56.1 & 79.4 & 65.7 & 35.7 & 74.0 & 61.5 & 46.0 & 67.1 \\
        & Ours & 70\% & \textbf{100.0} & \textbf{98.0} & \textbf{100.0} & \textbf{96.5} & \textbf{83.0} & \textbf{48.0} & \textbf{73.1} & \textbf{95.4} & \textbf{81.1} & \textbf{47.1} & \textbf{78.0} & \textbf{66.5} & \textbf{47.0} & \textbf{78.0} \\
        \midrule
          & Original & - & 100.0 & 99.5 & 66.0 & 92.5 & 56.0 & 8.5 & 63.1 & 90.5 & 81.2 & 36.5 & 57.2 & 43.5 & 34.5 & 63.6 \\
        128K & ThinK & 70\% & 97.0 & 93.5 & 93.5 & 91.5 & 39.5 & 8.0 & 55.4 & 72.8 & 55.8 & \textbf{36.1} & \textbf{61.3} & \textbf{46.5} & \textbf{34.5} & 60.4 \\
        & Ours & 70\% & \textbf{100.0} & \textbf{99.5} & \textbf{96.5} & \textbf{93.0} & \textbf{54.0} & \textbf{17.5} & \textbf{66.6} & \textbf{85.6} & \textbf{60.9} & 29.0 & 59.0 & 43.5 & 33.0 & \textbf{64.5} \\
        \bottomrule
    \end{tabular}}
    \caption{Performance of Llama-3.1-8B-Instruct and Qwen2.5-7B-Instruct on RULER. 200 samples are taken for each task.}
    \label{tab:ruler_full2}
\end{table*}

\begin{table*}[t]
    \centering
    \setlength{\tabcolsep}{1.8mm}
    \resizebox{2\columnwidth}{!}{
    \begin{tabular}{c|lc|cccccc|c}
        \toprule
        & & & \multicolumn{7}{c}{\textbf{8K Medium}}  \\ 
        & Method & \textbf{Pruning ratio} & \textbf{op=2} & \textbf{op=4} & \textbf{op=6} & \textbf{op=8} & \textbf{op=10} & \textbf{op=12} & \textbf{AUC} \\
        \midrule
          & Original & - & 0.1429 & 0.1389 & 0.1508 & 0.1349 & 0.0516 & 0.0278 & 0.5615 \\
        Llama-3.1-8B-Instruct & ThinK & 70\% & \textbf{0.1825} & 0.0556 & 0.0675 & 0.0159 & 0.0278 & 0.0079 & 0.2620 \\
        & Ours & 70\% & 0.1746 & \textbf{0.1944} & \textbf{0.2183} & \textbf{0.1389} & \textbf{0.0437} & \textbf{0.0397} & \textbf{0.7024} \\
        \midrule
          & Original & - & 0.2857 & 0.3651 & 0.2619 & 0.1865 & 0.0952 & 0.0556 & 1.0793 \\
        Qwen2.5-7B-Instruct & ThinK & 70\% & \textbf{0.4405} & \textbf{0.3532} & 0.1984 & 0.1190 & 0.0437 & 0.0437 & 0.9564   \\
        & Ours & 70\% & 0.3770 & 0.3333 & \textbf{0.2222} & \textbf{0.1587} & \textbf{0.1310} & \textbf{0.0476} &\textbf{ 1.0575} \\
        \bottomrule
    \end{tabular}}
    \resizebox{2\columnwidth}{!}{
    \begin{tabular}{c|lc|cccccc|c}
        & & & \multicolumn{7}{c}{\textbf{8K Hard}} \\ 
        & Method & \textbf{Pruning ratio} & \textbf{op=2} & \textbf{op=4} & \textbf{op=6} & \textbf{op=8} & \textbf{op=10} & \textbf{op=12} & \textbf{AUC} \\
        \midrule
          & Original & - & 0.4127 & 0.1865 & 0.2540 & 0.1905 & 0.1706 & 0.1190 & 1.0675  \\
        Llama-3.1-8B-Instruct & ThinK & 70\% & 0.0833 & 0.0833 & 0.0595 & 0.0198 & 0.0595 & 0.0357 & 0.2816  \\
        & Ours & 70\% & \textbf{0.4762} & \textbf{0.1944} & \textbf{0.2659} & \textbf{0.2103} & \textbf{0.1706} & \textbf{0.1270} & \textbf{1.1428}  \\
        \midrule
          & Original & - & 0.4643 & 0.2817 & 0.2262 & 0.1032 & 0.1151 & 0.1032 & 1.010  \\
        Qwen2.5-7B-Instruct & ThinK & 70\% & \textbf{0.3492} & 0.1944 & 0.1746 & 0.0913 & \textbf{0.0913} & 0.0635 & 0.7580  \\
        & Ours & 70\% & 0.3373 & \textbf{0.3532} & \textbf{0.2381} & \textbf{0.1389} & 0.0833 & \textbf{0.0952} & \textbf{1.0298}  \\
        \bottomrule
    \end{tabular}}

    \resizebox{2\columnwidth}{!}{
    \begin{tabular}{c|lc|cccccc|c}
        & & & \multicolumn{7}{c}{\textbf{16K Medium}} \\ 
        & Method & \textbf{Pruning ratio} & \textbf{op=2} & \textbf{op=4} & \textbf{op=6} & \textbf{op=8} & \textbf{op=10} & \textbf{op=12} & \textbf{AUC}  \\
        \midrule
          & Original & - & 0.0913 & 0.0675 & 0.1270 & 0.0833 & 0.0159 & 0.0159 & 0.3473  \\
        Llama-3.1-8B-Instruct & ThinK & 70\% & 0.1389 & 0.0397 & 0.0357 & 0.0159 & 0.0040 & 0.0119 & 0.1707 \\
        & Ours & 70\% & \textbf{0.1468} & \textbf{0.1071} & \textbf{0.1706} & \textbf{0.0913} & \textbf{0.0317} &\textbf{ 0.0317} & \textbf{0.4900} \\
        \midrule
          & Original & - & 0.2738 & 0.2937 & 0.2381 & 0.1429 & 0.0833 & 0.0556 & 0.9227  \\
        Qwen2.5-7B-Instruct & ThinK & 70\% & 0.4008 & \textbf{0.2778} & 0.1429 & 0.0833 & 0.0397 & 0.0119 & 0.7501  \\
        & Ours & 70\% & \textbf{0.4365} & 0.2302 & \textbf{0.1825} & \textbf{0.1468} & \textbf{0.0437} & \textbf{0.0714} & \textbf{0.8572}  \\
        \bottomrule
    \end{tabular}}
    \resizebox{2\columnwidth}{!}{
    \begin{tabular}{c|lc|cccccc|c}
        & & & \multicolumn{7}{c}{\textbf{16K Hard}} \\ 
        & Method & \textbf{Pruning ratio} & \textbf{op=2} & \textbf{op=4} & \textbf{op=6} & \textbf{op=8} & \textbf{op=10} & \textbf{op=12} & \textbf{AUC} \\
        \midrule
          & Original & - & 0.3214 & 0.0992 & 0.1429 & 0.1190 & 0.1071 & 0.0476 & 0.6527 \\
        Llama-3.1-8B-Instruct & ThinK & 70\% & 0.1032 & 0.0159 & 0.0198 & 0.0238 & 0.0159 & 0.0198 & 0.1369 \\
        & Ours & 70\% & \textbf{0.2857} & \textbf{0.1230} & \textbf{0.1349} & \textbf{0.1310} & \textbf{0.0833} & \textbf{0.0754} & \textbf{0.6528} \\
        \midrule
          & Original & - & 0.4722 & 0.2619 & 0.1706 & 0.1389 & 0.0873 & 0.0635 & 0.9265  \\
        Qwen2.5-7B-Instruct & ThinK & 70\% & 0.3770 & 0.1786 & 0.1429 & 0.0754 & 0.0635 & 0.0873 & 0.6926 \\
        & Ours & 70\% & \textbf{0.4246} & \textbf{0.2579} & \textbf{0.1548} & \textbf{0.0873} & \textbf{0.0913} & \textbf{0.0952} & \textbf{0.8512} \\
    \end{tabular}}

    \resizebox{2\columnwidth}{!}{
    \begin{tabular}{c|lc|cccccc|c}
        \toprule
        & & & \multicolumn{7}{c}{\textbf{32K Medium}} \\ 
        & Method & \textbf{Pruning ratio} & \textbf{op=2} & \textbf{op=4} & \textbf{op=6} & \textbf{op=8} & \textbf{op=10} & \textbf{op=12} & \textbf{AUC} \\
        \midrule
          & Original & - & 0.0873 & 0.0675 & 0.1151 & 0.0675 & 0.004 & 0 & 0.2978  \\
        Llama-3.1-8B-Instruct & ThinK & 70\% & 0.0913 & 0.0437 & 0.0437 & 0.0198 & 0 & 0 & 0.1529  \\
        & Ours & 70\% & \textbf{0.1190} & \textbf{0.0992} & \textbf{0.1429} & \textbf{0.0754} & \textbf{0.0198} & \textbf{0.0079} & \textbf{0.4007}  \\
        \midrule
          & Original & - & 0.3016 & 0.3730 & 0.2778 & 0.1468 & 0.0794 & 0.0714 & 1.0635  \\
        Qwen2.5-7B-Instruct & ThinK & 70\% & \textbf{0.4206} & 0.2183 & 0.1627 & 0.0913 & \textbf{0.0476} & \textbf{0.0437} & 0.7520  \\
        & Ours & 70\% & 0.3294 & \textbf{0.2381} & \textbf{0.2063} & \textbf{0.0992} & 0.0397 & 0.0317 & \textbf{0.7639}  \\
        \bottomrule
    \end{tabular}}
    \resizebox{2\columnwidth}{!}{
    \begin{tabular}{c|lc|cccccc|cc}
        & & & \multicolumn{7}{c}{\textbf{32K Hard}} \\ 
        & Method & \textbf{Pruning ratio} & \textbf{op=2} & \textbf{op=4} & \textbf{op=6} & \textbf{op=8} & \textbf{op=10} & \textbf{op=12} & \textbf{AUC} \\
        \midrule
          & Original & - & 0.1984 & 0.1190 & 0.1111 & 0.0675 & 0.0437 & 0.0437 & 0.4624 \\
        Llama-3.1-8B-Instruct & ThinK & 70\% & 0.0675 & 0.0278 & 0.0119 & 0.0119 & 0.0238 & 0.0278 & 0.1231  \\
        & Ours & 70\% & \textbf{0.2579} & \textbf{0.1468} & \textbf{0.0794} & \textbf{0.0714} & \textbf{0.0516} & \textbf{0.0516} & \textbf{0.5040}  \\
        \midrule
          & Original & - & 0.4603 & 0.2579 & 0.1468 & 0.1151 & 0.1032 & 0.0556 & 0.8810 \\
        Qwen2.5-7B-Instruct & ThinK & 70\% & \textbf{0.3929} & 0.1429 & 0.1429 & 0.0714 & 0.0635 & \textbf{0.0476} & 0.6409  \\
        & Ours & 70\% & \textbf{0.3929} & \textbf{0.1905} & \textbf{0.1587} & \textbf{0.1071} & \textbf{0.0675} & 0.0437 & \textbf{0.7421}  \\
        \bottomrule
    \end{tabular}}
    \caption{Performance of different methods and models on GSM-Infinite, 256 samples are taken for each op. Generation temperature is set to 0.}
    \label{tab:gsm_detail_2}
\end{table*}

\section{Full Evaluation Results on GSM-Infinite}
\label{sec:appendix}

We evaluate Medium and Hard tasks in GSM-Infinite across input lengths ranging from 8K to 32K. Full results are shown in Table \ref{tab:gsm_detail_2}.

\section{Additional Experimental Results} 

\label{sec:static}

Table \ref{tab:static-norm}, Table \ref{tab:ortho_rst} and Table \ref{tab:ablation_budget} provide other supplementary results. 

Specifically, Table \ref{tab:static-norm} evaluated static norm-based selection method's performance on RULER, suggesting that a static channel pruning strategy could achieve comparable performance with dynamic methods such as ThinK. The implementation of the method is illustrated in Section \ref{sec:motiv}.

Table \ref{tab:ortho_rst} verified that LeanK is orthogonal with token pruning (Quest), head pruning (DuoAttention) and quantization (KIVI) methods. For Quest, block size is 64 and token budget is 1024. For DuoAttention, 50\% heads are pruned. KIVI are tested with 2 bits for both K and V cache, group size 32 and residual length 128.

Table \ref{tab:ablation_budget} conducted ablation experiments to justify our design choices, showing that both (1) a more fine-grained head-wise budget allocation and (2) a learning-based channel-wise importance score that does not rely on channel's magnitude contribute to the effectiveness of our method. 

\section{Choice of Hyperparameter Lambda}

The hyperparameter $\lambda$ is multiplied to the regularization loss (L1 norm of scaling factor $\alpha$) in training Stage1, with the total loss defined as: $L_{1st} = L_{dist} + \lambda L_{reg}$. 

We trained models with varying $\lambda$ values on Llama-3.1-8B-Instruct and evaluated performance on the RULER 32K benchmark.

\begin{table}[H]
    \centering
    \small
    \setlength{\tabcolsep}{1.8mm}
    \resizebox{1\columnwidth}{!}{
    \begin{tabular}{c|lc|c}
        \toprule
        & Method & Ratio & Acc \\
        \midrule
        & Original & - & 86.0 \\
        & $\lambda$ = 0.04 & 70\% & 84.4 \\
        Llama-3.1-8B-Instruct & $\lambda$ = 0.06 & 70\% & \textbf{85.8} \\
        & $\lambda$ = 0.08 & 70\% & 85.5 \\
        & $\lambda$ = 0.10 & 70\% & 85.1 \\
        \bottomrule
    \end{tabular}
    }
    \caption{RULER 32K performance with different choices of training hyperparameter $\lambda$ on Llama-3.1-8B-Instruct. }
    \label{tab:choice_of_lambda}
\end{table}

Results in Table\ref{tab:choice_of_lambda} show that LeanK is robust to a wide range of $\lambda$ values (0.04 to 0.10), achieving the best performance around $\lambda=0.06$. Setting $\lambda$ too small (e.g., 0.04) under-regularizes $\alpha$, while excessively large $\lambda$ values (e.g., 0.10) can lead to less accurate learning during Stage 1, slightly harming the final results. 

\section{Choice of Training Task}

LeanK is robust to the choice of training task. Specifically, we trained Qwen2.5-7B-Instruct using DuoAttention[1]’s task and evaluated performance on RULER 16K. Results in Table \ref{tab:choice_of_task} verified that LeanK could still effectively learn channel importance with different tasks (with training hyperparameters changed accordingly), suggesting that LeanK's effectiveness is not sensitive to specific training tasks.

\begin{table}[H]
    \centering
    \small
    \setlength{\tabcolsep}{1.8mm}
    \resizebox{1\columnwidth}{!}{
    \begin{tabular}{c|lc|c}
        \toprule
        & Method & Ratio & Acc \\
        \midrule
        & Original & - & 90.8 \\
        Qwen2.5-7B-Instruct & Our Task & 70\% & \textbf{90.6} \\
        & DuoAttn Task & 70\% & 90.2 \\
        \bottomrule
    \end{tabular}
    }
    \caption{LeanK's performance on RULER 16K with different choices of training task $\lambda$ on Qwen2.5-7B-Instruct. }
    \label{tab:choice_of_task}
\end{table}

\section{Comparison with Static Channel Pruning Baseline}

We evaluated a static norm-based channel pruning baseline on Llama-3.1-8B-Instruct using the RULER benchmark. The static mask was generated by averaging channel norm distribution (as described in Section 2.2) across 100 NIAH\_multikey3 sequences with 64K context length. Results are shown in Table \ref{tab:static_baseline}.

\begin{table}[H]
    \centering
    \small
    \setlength{\tabcolsep}{1.8mm}
    \resizebox{1\columnwidth}{!}{
    \begin{tabular}{cc|cccccc|c}
        \toprule
        Method & Ratio & 4K & 8K & 16K & 32K & 64K & 128K & Avg. \\
        \midrule
        Original & - & 95.1 & 93.1 & 90.2 & 86.0 & 84.4 & 73.5 & 87.1 \\
        ThinK & 70\% & 58.3 & 39.2 & 37.1 & 36.4 & 35.3 & 40.0 & 39.4 \\
        Static & 70\% & 68.7 & 57.5 & 54.8 & 55.9 & 54.0 & 55.5 & 57.7 \\
        LeanK & 70\% & 95.3 & 93.4 & 88.8 & 85.8 & 84.1 & 73.2 & \textbf{86.8} \\
        \bottomrule
    \end{tabular}
    }
    \caption{Different method's performance with Llama-3.1-8B-Instruct on the RULER dataset.}
    \label{tab:static_baseline}
\end{table}

The results show that averaging channel norms across multiple inputs yields more effective pruning than dynamic methods like ThinK. However, there remains a substantial performance gap compared to LeanK (i.e., Learned Pattern > Static Pattern > Dynamic Pattern) . This further supports both the static nature of channel importance and the effectiveness of LeanK’s design. 

\section{Analysis of V-Cache Pruning Conditions}

We analyzed both the distribution and characteristics of the completely pruned heads:

1) In terms of \textbf{Distribution}, these heads are distributed across all layers, with a higher concentration in the shallow layers (e.g., in Qwen, 43\% of them appear in the first 5 layers). 

2) Speaking of \textbf{Characteristics}, We examined the channel frequency distribution of these heads (as described in Section\ref{sec:high-frequency-sensitivity}). These heads exhibit high \textit{high-frequency ratio} $w_{hf}$ and relatively large norms on high-frequency channels. This indicates that these heads mainly rely on local context and patterns for next-token prediction, while contributing less to semantic extraction from the context.

\section{Comparison with SnapKV}

We compared LeanK with SnapKV, a token-level KV cache pruning method, with results shown in \ref{tab:snapkv_comparison}. 
Under the same overall KV cache reduction ratio (44\% for Llama, 43\% for Qwen), LeanK achieves higher average performance. SnapKV underperforms LeanK on complex KV retrieval tasks such as NIAH\_multikey3. 

\begin{table*}[t]
    \centering
    \setlength{\tabcolsep}{1.8mm}
    \resizebox{2\columnwidth}{!}{
    \begin{tabular}{c|c|ccccccccccccc|c}
        \toprule
        Model & Method & \textbf{niah\_s1} & \textbf{niah\_s2} & \textbf{niah\_s3} & \textbf{niah\_mk1} & \textbf{niah\_mk2} & \textbf{niah\_mk3} & \textbf{niah\_mv} & \textbf{niah\_mq} & \textbf{vt} & \textbf{cwe} & \textbf{fwe} & \textbf{qa\_1} & \textbf{qa\_2} & Avg. \\
        \midrule
          Llama-3.1-8B & SnapKV & 100 & 100 & 100 & 99.5 & 99.0 & 89.5 & 99.1 & 99.4 & 96.9 & 5.5 & 92.5 & 76.5 & 49.5 & 85.2 \\
          & LeanK & 100 & 100 & 100 & 99.5 & 99.5 & 100.0 & 98.4 & 98.5 & 97.0 & 1.2 & 93.3 & 76.0 & 51.5 & \textbf{85.8} \\
        Qwen2.5-7B & SnapKV & 100 & 100 & 100 & 99.0 & 93.5 & 79.0 & 90.6 & 98.9 & 97.8 & 72.5 & 88.2 & 67.0 & 55.0 & 87.8 \\
         & LeanK & 100 & 100 & 100 & 99.0 & 97.5 & 93.5 & 85.3 & 98.9 & 96.0 & 67.8 & 90.5 & 65.5 & 56.0 & \textbf{88.5} \\
        \bottomrule
    \end{tabular}}
    \caption{Performance of LeanK and SnapKV on RULER 32K under the same pruning ratio. }
    \label{tab:snapkv_comparison}
\end{table*}

\begin{table*}[t]
    \centering
    \setlength{\tabcolsep}{1.8mm}
    \resizebox{2\columnwidth}{!}{
    \begin{tabular}{c|lc|ccccccccccccc|c}
        \toprule
        & & & \multicolumn{13}{c}{\textbf{Llama-3.1-8B-Instruct}} &  \\ 
        & Method & \textbf{Ratio} & \textbf{niah\_s1} & \textbf{niah\_s2} & \textbf{niah\_s3} & \textbf{niah\_mk1} & \textbf{niah\_mk2} & \textbf{niah\_mk3} & \textbf{niah\_mv} & \textbf{niah\_mq} & \textbf{vt} & \textbf{cwe} & \textbf{fwe} & \textbf{qa\_1} & \textbf{qa\_2} & Avg. \\
        \midrule
          & Original & - & 100.0 & 100.0 & 100.0 & 100.0 & 97.0 & 99.0 & 97.0 & 99.1 & 94.5 & 0.1 & 85.3 & 75.5 & 49.5 & 84.4 \\
        RULER & Doule Sparsity & 60\% & 45.5 & 37.0 & 37.5 & 35.0  & 36.0 & 1.0 & 8.9 & 20.9 & 14.8 & 0.1 & 71.5 & 66.0 & 45.0 & 29.9 \\
        64K & Doule Sparsity & 70\% & 3.0 & 5.5 & 3.5 & 9.0 & 3.0 & 1.0 & 3.5 & 5.6 & 1.1 & 0.1 & 68.8 & 54.0 & 37.0 & 14.0 \\
        & Ours & 70\% & 100.0 & 100.0 & 100.0 & 99.0 & 97.0 & 99.0 & 95.6 & 98.5 & 94.3 & 0.1 & 85.3 & 75.5 & 49.0 & 84.1 \\
        \bottomrule
    \end{tabular}}
    \caption{Full comparison results comparing our method with Double Sparsity. }
    \label{tab:ds_rst}
\end{table*}

\begin{table*}[t]
    \centering
    \setlength{\tabcolsep}{1.8mm}
    \resizebox{2\columnwidth}{!}{
    \begin{tabular}{l|cccccccccccccc|c}
        \toprule
         & & \multicolumn{12}{c}{\textbf{Qwen2.5-7B-Instruct}} & \\ 
         Method & \textbf{Ratio} & \textbf{niah\_s1} & \textbf{niah\_s2} & \textbf{niah\_s3} & \textbf{niah\_mk1} & \textbf{niah\_mk2} & \textbf{niah\_mk3} & \textbf{niah\_mv} & \textbf{niah\_mq} & \textbf{vt} & \textbf{cwe} & \textbf{fwe} & \textbf{qa\_1} & \textbf{qa\_2} & Avg. \\
        \midrule
           Original & - & 100.0 & 98.0 & 98.0 & 95.5 & 82.5 & 47.5 & 82.8 & 97.4 & 95.3 & 55.4 & 82.5 & 69.5 & 44.5 & 80.7 \\
        ThinK (Dynamic norm) & 60\% & 97.0 & 95.5 & 100.0 & 98.5 & 94.0 & 98.0 & 90.1 & 97.8 & 72.0 & 0.1 & 80.33 & 76.0 & 48.0 & 80.6 \\
          Static norm & 60\% & 96.5 & 96.5 & 100.0 & 96.0 & 95.0 & 99.5 & 86.4 & 96.1 & 91.5 & 0.1 & 81.3 & 74.0 & 47.5 & 81.6 \\
        \bottomrule
    \end{tabular}}
    \caption{Static norm-based selection. Tested on RULER 64K, with 200 samples from each subtask. }
    \label{tab:static-norm}
\end{table*}

\begin{table*}[t]
    \centering
    \setlength{\tabcolsep}{1.8mm}
    \resizebox{2\columnwidth}{!}{
    \begin{tabular}{c|lc|ccccccccccccc|c}
        \toprule
        & & & \multicolumn{12}{c}{\textbf{Llama-3.1-8B-Instruct}} & \\ 
        & Method & \textbf{Ratio} & \textbf{niah\_s1} & \textbf{niah\_s2} & \textbf{niah\_s3} & \textbf{niah\_mk1} & \textbf{niah\_mk2} & \textbf{niah\_mk3} & \textbf{niah\_mv} & \textbf{niah\_mq} & \textbf{vt} & \textbf{cwe} & \textbf{fwe} & \textbf{qa\_1} & \textbf{qa\_2} & Avg. \\
        \midrule
          & Original & - &  100.0 & 100.0 & 100.0 & 100.0 & 97.0 & 99.0 & 97.0 & 99.1 & 94.5 & 0.1 & 85.3 & 75.5 & 49.5 & 84.4 \\
         & DuoAttn & 50\% & 100.0 & 100.0 & 100.0 & 99.0 & 96.5 & 99.0 & 95.8 & 99.3 & 91.2 & 0.1 & 84.5 & 76.0 & 50.0 & 84.0 \\
        64K & DuoAttn + Ours & 80\% & 100.0 & 100.0 & 100.0 & 99.0 & 96.0 & 99.0 & 95.0 & 99.4 & 89.8 & 0.1 & 83.7 & 76.0 & 48.0 & 83.5 \\
         & Quest & - & 100.0 & 98.5 & 82.0 & 98.5 & 84.0 & 5.0 & 93.3 & 95.1 & 83.2 & 0.6 & 83.7 & 72.5 & 45.0 & 72.4 \\
         & Quest + Ours & 70\% & 100.0 & 100.0 & 99.0 & 99.0 & 84.0 & 11.5 & 93.0 & 96.9 & 87.6 & 0.05 & 87.3 & 74.0 & 44.5 & 75.1 \\
    \midrule
        & & & \multicolumn{12}{c}{\textbf{Llama-3.1-8B-Instruct}} & \\ 
        & Method & \textbf{Ratio} & \textbf{niah\_s1} & \textbf{niah\_s2} & \textbf{niah\_s3} & \textbf{niah\_mk1} & \textbf{niah\_mk2} & \textbf{niah\_mk3} & \textbf{niah\_mv} & \textbf{niah\_mq} & \textbf{vt} & \textbf{cwe} & \textbf{fwe} & \textbf{qa\_1} & \textbf{qa\_2} & Avg. \\
        \midrule
          & Original & - & 100.0 & 100.0 & 100.0 & 100.0 & 99.0 & 100.0 & 98.9 & 99.4 & 97.6 & 2.7 & 93.3 & 76.0 & 51.5 & 86.0   \\
        32K & KIVI & - & 100.0 & 100.0 & 100.0 & 98.5 & 96.5 & 95.3 & 97.5 & 98.8 & 91.4 & 5.8 & 93.5 & 75.5 & 48.0 & 84.7 \\
        & KIVI + Ours & 70\% & 100.0 & 99.5 & 100.0 & 99.0 & 98.0 & 91.5 & 97.1 & 98.4 & 92.3 & 1.0 & 96.3 & 73.5 & 47.5 & 84.2 \\
        \bottomrule
    \end{tabular}}
    \caption{LeanK applied on top of other pruning methods. Tested on RULER 64K (32K for KIVI to avoid OOM), with 200 samples from each subtask. }
    \label{tab:ortho_rst}
\end{table*}

\begin{table*}[t]
    \centering
    \setlength{\tabcolsep}{1.8mm}
    \resizebox{2\columnwidth}{!}{
    \begin{tabular}{l|cccccccccccccc|c}
        \toprule
         & & \multicolumn{12}{c}{\textbf{Qwen2.5-7B-Instruct}} & \\ 
         Method & \textbf{Ratio} & \textbf{niah\_s1} & \textbf{niah\_s2} & \textbf{niah\_s3} & \textbf{niah\_mk1} & \textbf{niah\_mk2} & \textbf{niah\_mk3} & \textbf{niah\_mv} & \textbf{niah\_mq} & \textbf{vt} & \textbf{cwe} & \textbf{fwe} & \textbf{qa\_1} & \textbf{qa\_2} & Avg. \\
        \midrule
           Original & - & 100.0 & 98.0 & 98.0 & 95.5 & 82.5 & 47.5 & 82.8 & 97.4 & 95.3 & 55.4 & 82.5 & 69.5 & 44.5 & 80.7 \\
         w/o 2nd stage & 70\% & 99.5 & 86.5 & 99.0 & 92.0 & 75.5 & 10.5 & 77.4 & 91.1 & 58.3 & 48.5 & 72.7 & 64.5 & 44.0 & 70.7 \\
          w/ 2nd stage & 70\% & 100.0 & 98.0 & 100.0 & 96.5 & 83.0 & 48.0 & 73.1 & 95.4 & 81.1 & 47.1 & 78.0 & 66.5 & 47.0 & 78.0 \\
        \bottomrule
    \end{tabular}}
    \caption{Necessity of the second stage of training. Tested on RULER 64K, with 200 samples from each subtask. }
    \label{tab:2ndstage_details}
\end{table*}

\begin{table*}
    \centering
    \setlength{\tabcolsep}{1.8mm}
    \resizebox{2\columnwidth}{!}{
    \begin{tabular}{l|cccccccccccccc|c}
        \toprule
        & & \multicolumn{12}{c}{\textbf{Llama-3.1-8B-Instruct}} & \\ 
         Method & \textbf{Ratio} & \textbf{niah\_s1} & \textbf{niah\_s2} & \textbf{niah\_s3} & \textbf{niah\_mk1} & \textbf{niah\_mk2} & \textbf{niah\_mk3} & \textbf{niah\_mv} & \textbf{niah\_mq} & \textbf{vt} & \textbf{cwe} & \textbf{fwe} & \textbf{qa\_1} & \textbf{qa\_2} & Avg. \\
        \midrule
           Original & - &  100.0 & 100.0 & 100.0 & 100.0 & 97.0 & 99.0 & 97.0 & 99.1 & 94.5 & 0.05 & 85.3 & 75.5 & 49.5 & 84.4 \\
          Uneven Dynamic & 70\% & 100.0 & 100.0 & 98.5 & 99.0 & 92.5 & 92.5 & 86.9 & 93.3 & 39.6 & 1.0 & 70.5 & 74.0 & 48.0 & 76.6 \\
          LeanK & 70\% & 100.0 & 100.0 & 100.0 & 99.0 & 97.0 & 99.0 & 95.6 & 98.5 & 94.3 & 0.05 & 85.3 & 75.5 & 49.0 & 84.1 \\
        \bottomrule
    \end{tabular}}
    \caption{Using our trained mask's budget allocation and ThinK's dynamic norm-based channel selection strategy for pruning. Tested on RULER 64K, with 200 samples from each subtask. }
    \label{tab:ablation_budget}
\end{table*}